\documentclass[10pt,journal,compsoc]{IEEEtran}
\ifCLASSOPTIONcompsoc
  \usepackage[nocompress]{cite}
\else
  \usepackage{cite}
\fi

\ifCLASSINFOpdf
\else
\fi

\usepackage{times}
\usepackage{epsfig}
\usepackage{graphicx}
\usepackage{amsmath}
\usepackage{amssymb}
\usepackage{mathrsfs}
\usepackage{multirow}
\usepackage{multicol}
\usepackage{lipsum}
\usepackage{float}
\usepackage{svg}
\usepackage{subcaption}
\usepackage{multicol}

\begin{document}
\title{Multi-modality Deep Restoration of \\ Extremely Compressed Face Videos}

\author{Xi~Zhang and
        Xiaolin~Wu,~\IEEEmembership{Fellow,~IEEE}
\IEEEcompsocitemizethanks{
\IEEEcompsocthanksitem X.~Zhang is with the Department of Electronic Engineering, Shanghai Jiao Tong University, Shanghai, 200204, China. \protect\\
E-mail: zhangxi\_19930818@sjtu.edu.cn
\IEEEcompsocthanksitem X.~Wu is with the Department of Electrical \& Computer Engineering, McMaster University, Hamilton, L8G 4K1, Ontario, Canada. \protect\\
Email: xwu@ece.mcmaster.ca}
\thanks{Manuscript received April 19, 2005; revised August 26, 2015.}}

%
%

\markboth{Manuscript submitted to IEEE TRANSACTIONS ON PATTERN ANALYSIS AND MACHINE INTELLIGENCE}
{Zhang \MakeLowercase{\textit{et al.}}: Multi-modality Deep Restoration of Extremely Compressed Face Videos}

%



\IEEEtitleabstractindextext{%
\begin{abstract}
Arguably the most common and salient object in daily video communications is the talking head, as encountered in social media, virtual classrooms, teleconferences, news broadcasting, talk shows, etc.  When communication bandwidth is limited by network congestions or cost effectiveness, compression artifacts in talking head videos are inevitable.  The resulting video quality degradation is highly visible and objectionable due to high acuity of human visual system to faces.  To solve this problem, we develop a multi-modality deep convolutional neural network method for restoring face videos that are aggressively compressed. The main innovation is a new DCNN architecture that incorporates known priors of multiple modalities: the video-synchronized speech signal and semantic elements of the compression code stream, including motion vectors, code partition map and quantization parameters. These priors strongly correlate with the latent video and hence they are able to enhance the capability of deep learning to remove compression artifacts.  Ample empirical evidences are presented to validate the superior performance of the proposed DCNN method on face videos over the existing state-of-the-art methods.

\end{abstract}

\begin{IEEEkeywords}
Multi-modality, video restoration, face videos, deep neural networks.
\end{IEEEkeywords}}

\maketitle

\IEEEdisplaynontitleabstractindextext

%
\IEEEpeerreviewmaketitle

\IEEEraisesectionheading{\section{Introduction}\label{sec:introduction}}

\IEEEPARstart{V}{ideo} contents create well over sixty percent of data traffic on the Internet, and this percentage is still steadily climbing as more and more of people's daily interactions are conducted on line.  Communication bandwidths and data storages are under constant pressures due to rapid expansion and ubiquity of on-line video applications.  As such, video compression has been and will continue to be an indispensable enabling technology in the modern digital world.  For most users, video files have to be compressed by one of the popular video compression methods (e.g., MPEG-4~\cite{mpeg4}, H.264~\cite{h264}, HEVC~\cite{h265}) to a sufficiently small size to achieve an acceptable level of cost effectiveness.
For high compression ratio or low bit rates, lossy video compression inevitably produces objectionable artifacts, such as blocking, blurring, ringing and jaggies.
Recently quite a few deep learning methods are proposed to remove video compression artifacts.  Compared with the pure end-to-end DCNN approach for video compression~\cite{dvc,videoRD}, the methods of compression artifacts removal~\cite{ DKFN,MFQE} have the operational advantage of being compatible with existing video compression standards, as they are essentially a post-processing step of restoring already-decoded videos by the standards. We call this CNN-based video restoration strategy deep video decompression.

This paper is concerned with deep face video decompression.
Talking heads are arguably the most common and salient object in daily Internet video communications. For examples, conversational faces are the focal centerpiece in social media, remote education, news media, teleconferences, Internet talk shows (TED and the alike), self media, etc.  If the network communication capacity is hard pressed, for instance, many large virtual but highly interactive classes are conducted by a school at the same time, achieving high reconstruction quality of faces in compressed videos is challenging but it is
crucial to satisfactory user experiences.  This is the practical motivation of our research.

Much to the advantage of video restoration algorithm designers, faces in conversation have very strong priors that can greatly reduce the solution space of the underlying inverse problem.  At very least, face is a known highly structured object; in some cases the algorithm even knows the particular person whose face is to restore.  Furthermore, the speech content of the speaking person is also available.  Physiologically, facial muscles, particularly those in the lips, shape the sound and air stream into speech.  This is why people can read lips, i.e., recognizing uttered words by watching the speaker's lips even without sound.  In addition to accompanying speech, the semantic elements and structural information of the video codec in question, such as motion vectors, motion residuals, coding block organization, quantization parameters, etc., also offer strong priors that can aid the video restoration task.


\begin{figure*}
  \centering
  \includegraphics[width=0.95\textwidth]{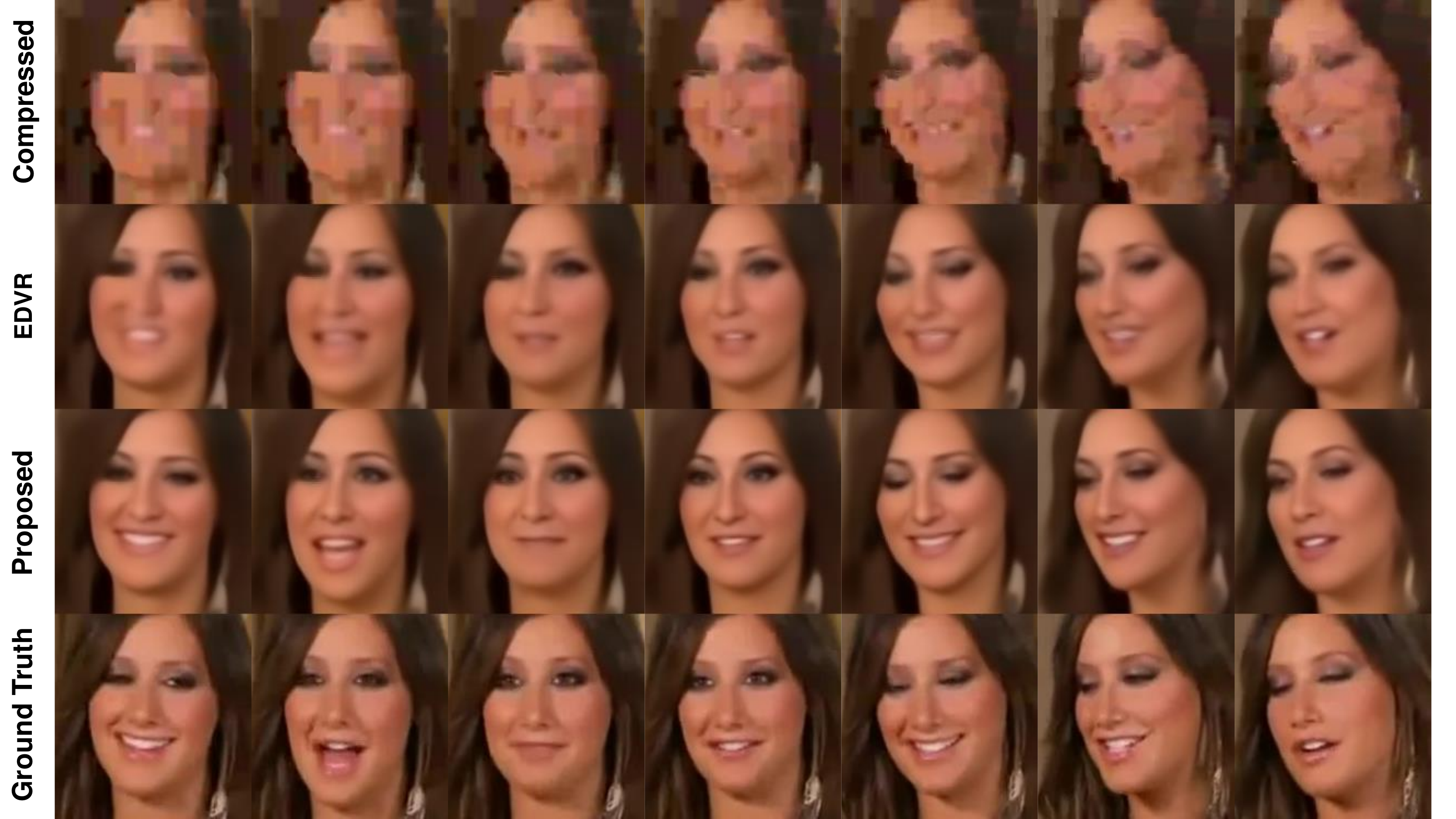}
  \caption{Visual comparisons of the proposed MDVD-Net and the stat-of-the-art method EDVR. MDVD-Net can produce more precise mouth shape, clearer teeth, sharper lips and muscle contours.}
  \label{demo}
\end{figure*}

In this work we design a novel neural network architecture, called Multi-modality Deep Video Decompression Network (MDVD-Net), to significantly improve the quality of aggressively compressed face videos (see Fig.~\ref{demo} for preview).  The success of the MDVD-Net depends on how effectively the network can incorporate and profit from the above priors of multiple modalities, which are largely overlooked by the existing methods.  The key technical developments include:
(1) A novel architecture that can exploit the high correlation of a person’s speech and her/his facial dynamics.
In addition to directly learning 2D features from speech, we use facial landmark points as intermediate representations and associate them with speech signals. Thanks to the low degree of freedoms for facial landmarks, our method can achieve good performance on moderately sized training datasets.
(2) A multi-frame alignment module guided by motion vectors which is obtained from code stream without any cost.
(3) A spatial attention module to dynamically fuse various priors from multiple modalities.
(4) A back projection module which refines the network output results by imposing both upper and lower bounds on the ground truth DCT coefficients of prediction residuals defined in video compression standards.






In summary, the major contributions of this research are the follows:
\vspace{-0.2cm}
\begin{itemize}
\item[1)] A baseline CNN method for deep decompression of talking head videos that outperforms existing CNN methods for the removal of compression artifacts, particularly at very low bit rates.
\item[2)] The MDVD-Net architecture design that exploits the priors of multiple modalities for added performance gains over the baseline method.
\item[3)] A systematic performance evaluation and analysis of the MDVD-Net methodology with the availability of priors of different strengths: the MDVD-Net trained for compressed videos of a particular known speaker with and without his/her voices, and for generic talking heads with and without the accompanying voices.
\end{itemize}

This research will have a lasting practical significance even minus the pressure of communication bandwidth exerted by mass social media in extensively networked virtual communities.  As the data volume of face-to-face conversations increases by order of magnitude from voice to video, backing up all conversational videos is unsustainable even for big social media service providers. The MDVD-Net technique allows such video contents to be archived in aggressively compressed form without the risk of fidelity loss, because compression defects can be repaired if the contents are ever recalled in the future.

This work significantly expands the technical scope and improves the performance of its earlier prototype presented at CVPR 2020 \cite{davd}.
The CNN architecture has been resigned to integrate and exploit useful priors of multiple modalities beyond the speech signal. Specifically, we incorporate into the CNN restoration process semantic elements of the video compression code stream, including motion vectors, code partition map and quantization parameters, and combine them with the speech prior.
Moreover, we exploit the statistical correlations between the speech and video modalities more thoroughly. Instead of directly learning features from speech, we use facial landmark points as intermediate representations and associate them with speech signals. The degree of freedoms for facial landmarks is in the order of tens (68 in our implementation), as opposed to millions of pixels in conventional 2D feature maps. This reduction allows the use of moderately sized datasets without performance loss.
The improved network model MDVD-Net outperforms not only the existing state-of-the-art methods, but also our previous model DAVD-Net \cite{davd}.

The rest of the paper is organized as follows.  After a brief review of related works in Section 2, we present, in Section 3, the justifications and details of our network design.  In Section 4, we describe the design of our experiments, explain the datasets used, and report our empirical findings.  The experiments demonstrate that the proposed MDVD-Net outperforms the existing state-of-the-art methods on videos of talking heads for compression artifact reduction. Section 5 concludes the paper.

\section{Related Work}
\textbf{Image compression artifact reduction.}
There is a large body of literature on removing compression artifacts in images~\cite{rw_foi, rw_zhang, rw_li, rw_chang, rw_dar, rw_liu, zhou2011, zhou2012, shu2017}. The majority of the studies on the subject focus on post-processing JPEG images to alleviate compression noises, apparently because JPEG is the most widely used lossy compression standard.

Inspired by successes of deep learning in image restoration, a number of CNN-based compression artifacts removal methods were developed~\cite{ARCNN, rw_svoboda, CAR_guo, CAR_galteri}.
Borrowing the CNN for super-resolution (SRCNN), Dong~\textit{et~al.}~\cite{ARCNN} proposed an artifact reduction CNN (ARCNN). The ARCNN has a three-layer structure: a feature extraction layer, a feature enhancement layer, and a reconstruction layer. This CNN structure is designed in the principle of sparse coding.  It was improved by Svoboda~\textit{et~al.}~\cite{rw_svoboda} who combined residual learning and symmetric weight initialization.
Guo~\textit{et~al.}~\cite{CAR_guo} and Galteri~\textit{et~al.}~\cite{CAR_galteri} proposed to reduce compression artifacts by Generative Adversarial Network (GAN), as GAN is able to generate sharper image details.
Zhang et al~\cite{calic,ultra} proposed to incorporate an $\ell_\infty$ fidelity criterion in the design of networks to protect small, distinctive structures in the framework of near-lossless image compression.

\textbf{Deep restoration of compressed videos.}
All above methods for image compression artifact reduction can be viewed as of single-frame approach to video restoration without using any temporal correlations between neighboring frames.
Yang~\textit{et~al.}~\cite{MFQE} introduced the first CNN-based multi-frame method for restoring compressed videos, which takes advantage of information in the neighboring frames.
Xue~\textit{et~al.}~\cite{TOFLOW} proposed a multi-task learning approach to jointly carry out motion estimation and a video restoration task.
He~\textit{et~al.}~\cite{partition} utilized the coding block information of the encoder structures
to guide the video decompression process.
Lu.~\textit{et~al.}~\cite{DKFN} modeled the video artifact reduction task as a Kalman filtering procedure and restored decoded frames through a deep Kalman filtering network. The main idea is to utilize the less noisy previously restored frames instead of directly decoded frames as temporal references.
Xu~\textit{et~al.}~\cite{nonlocal} introduced an non-local strategy in ConvLSTM to trace the spatiotemporal dependency in a video sequence, and achieved the state-of-the-art performance.
Zhang~\textit{et~al.}~\cite{davd, guo2020} proposed to restore talking-head videos using information from the audio stream and structural information given by the video encoder.

\textbf{End-to-end deep video compression.}
In recent years, a number of papers were published on the pure
end-to-end neural network approach to video compression, in which all coding operations, including motion estimation and compensation, residual compression, motion vector compression, quantization, and bit rate estimation, are carried out by the neural network.  The encoder and decoder are jointly optimized in rate-distortion trade-off via a single loss function
~\cite{dvc,learned,neural,scale,mlvc,improving}. However, as of now, all pure end-to-end neural video compression methods have prohibitively high computational complexity, far from being applicable in practice.  Moreover, pure neural video compression is completely incompatible with the deeply entrenched video/image compression standards in industrial and commercial worlds, whereas the methods researched in this paper are.







\textbf{Face restoration}
Due to the importance of and advanced research on face images,
many papers have been published on deep learning based face image restoration~\cite{zhu2016deep,cao_attention,fsrnet,chrysos,kim2019progressive,huang2017wavelet,xu2017learning,yu2018face,yu2018super}.
To recover the facial features in fine details, it is common to exploit the known priors on faces in the CNN restoration process. Shen~\textit{et~al.}~\cite{shen2018deep} proposed to learn a global semantic face prior and use it as network input to impose local structure on the output. Similarly, Xu~\textit{et~al.}~\cite{yu2018face} used a multi-task model to predict the facial components heatmaps for incorporating structure information. Chen~\textit{et~al.}~\cite{fsrnet} proposed to learn the facial geometry priors (i.e., landmarks heatmaps and parsing maps) for better recovery. Yu~\textit{et~al.}~\cite{yu2018super} developed a facial attribute-embedded network by incorporating face attributes vector in the LR feature space. Kim~\textit{et~al.}~\cite{kim2019progressive} adopted a progressive scheme to generate successive higher resolution outputs and proposed a facial attention loss on landmarks to constrain the structure of reconstruction.
Some other works use an additional face image of the same identity to guide the face restoration process, lending details to the restored face image~\cite{li2018learning}.



\textbf{Audio-driven talking head animation.}
Another line of research related to this work is
audio-driven talking head animation.
In 1999, Brand~\cite{voice} pioneered the work of Voice Puppetry to generate full facial animation from an audio track.
Suwajanakorn~\textit{et~al.}~\cite{obama} proposed an interesting technique to automatically edit a video of a given speaker with accurate lip synchronization guided by his own audio in a different speech.
This work has spawned in recent years a number of variant methods on the task
~\cite{said,temporal,viseme,eskimez2018,eskimez2019,x2face,talking,chen2019hierarchical,zakharov2019few,thies2020neural,zhou2020makelttalk,zhou2021pose}.

\begin{figure*}[t]
  \centering
  \includegraphics[width=\linewidth]{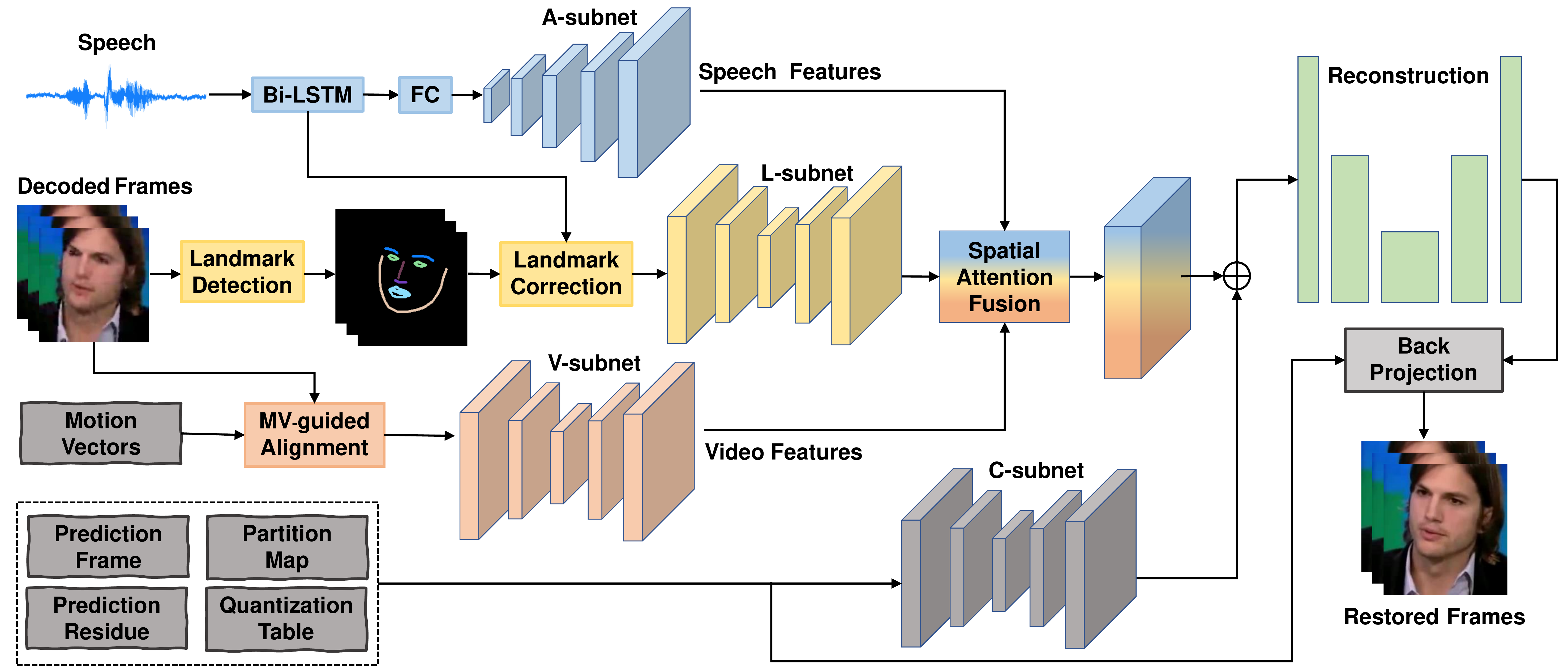}
  \caption{The framework of the proposed Multi-modality Deep Video Decompression Network (MDVD-Net). It consists of four branches, for speech, video, landmarks and codec information respectively.  }
  \label{overall}
\end{figure*}

\section{Methodology}
\subsection{Overview}
Given an original video sequence $\{X_t | t=0,1,2...\}$,  $\{X_t\}$ is to be compressed by a video compression standard (e.g. H.264/265) that removes spatial and temporal redundancy to gain efficiency in transmission and storage.
Then the compressed video will be decompressed by users to obtain a decoded video sequence,
denoted by $\{Y_t | t=0,1,2...\}$.

In the deep decompression task, the aim is to compute a refined reconstruction $\hat{X}_t$ from a decoded frame $Y_t$ by maximally removing compression artifacts in $Y_t$. In order to utilize the temporal information, most existing methods take the current decoded frame and neighboring frames as input and output a restored current frame, that is:
\begin{align}
\centering
\hat{X}_t = G( \mathcal{Y}_{t \pm n} )
\end{align}
where $\mathcal{Y}_{t \pm n} = \{Y_{t-n},...,Y_{t+n}\}$ denotes a consecutive $(2n+1)$ compressed frames and $G$ is the CNN to be optimized.  For the task of restoring compressed talking heads videos, the pertaining speech signal is a useful piece of information due to the strong correlation between speech and facial movements.  To take advantage of the speech, we reformulate the video reconstruction problem as:
\begin{align}
\centering
\hat{X}_t = G( \mathcal{Y}_{t \pm n},\ \mathcal{A}_{t \pm m} )
\end{align}
where $\mathcal{A}_{t \pm m} = \{A_{t-m},...,A_{t+m}\}$ is the speech signal temporally centered at $A_t$.

The overall architecture of the proposed MDVD-Net is shown in Fig.~\ref{overall}.
It consists of four branches, for speech, video, face landmarks and codec information respectively.
In the speech processing branch, we apply bidirectional LSTM to extract speech features, and feed them to a generation network (A-subnet) that produces a cluster of 2-D feature maps in preparation for being combined with other modalities.
In the video processing branch, after motion vector guided alignment and fusion of neighboring decoded frames, we design a V-subnet to extract features of the aligned frames.
In the landmarks branch, we firstly detect face landmarks from the decoded frames and then use speech to refine the detected landmarks, especially the landmarks around the mouth and eyes. After landmark correction, we adopt a L-subnet to extract deep features from the corrected facial landmarks.
In the codec information branch, the semantic elements and structural information of the video codec, such as prediction frame, prediction residue and partition map, etc., are fed into an C-subnet to extract features. Next, we design a spatial attention fusion module to dynamically fuse the speech, video and landmark features. After that, the above fused features are concatenated with the codec information features and fed into a reconstruction module.
Finally, before output, the reconstructed video is further improved by a back projection module that constrains the solution space by the quantization boundaries in the transform domain of the video compression standard.

Next, we detail the individual components of the proposed MDVD-Net.


\begin{figure*}[t]
  \centering
  \includegraphics[width=\linewidth]{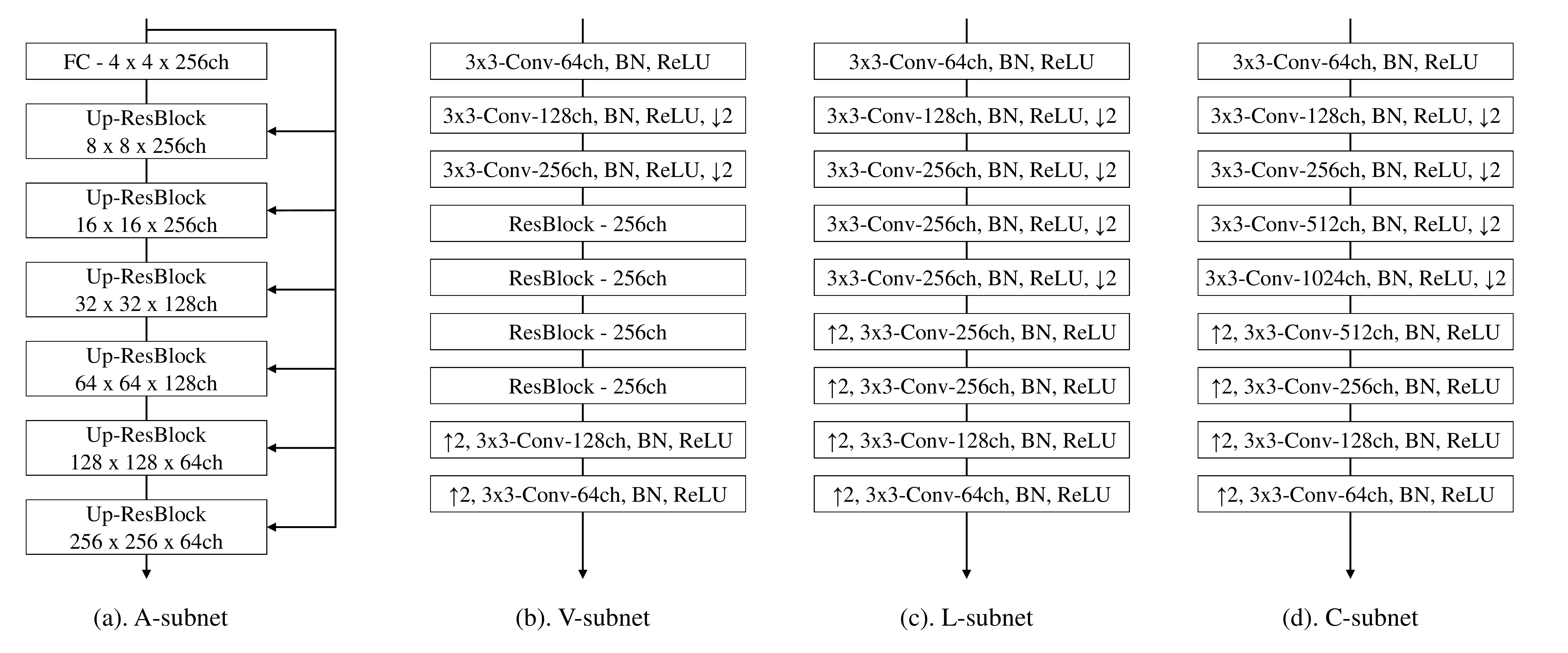}
  \caption{Architectures of individual subnets in our model. (a) is the architecture of A-subnet used to produce 2-D feature maps from speech signal in preparation for being combined with other modalities; (b) is the architecture of V-subnet used to  extract features of the aligned video frames; (c) is the architecture of L-subnet used to extract deep features from the corrected facial landmarks; (d) is the architecture of C-subnet used to extract features from the video codec information.}
  \label{subnets}
\end{figure*}

\subsection{Speech feature extraction}
In our design the speech signal is represented by the standard Mel-frequency cepstral coefficients (MFCC)~\cite{MFC,MFCC}.
When a person talks, at each time instance $t$, the facial image, particularly in parts around the mouth, depends not only on the current speech frame $A_t$ but also on previous and future speech frames.
For this reason, the network takes a consecutive speech feature sequence $\{A_{t-m},...,A_{t+m}\}$ as input in order to benefit from higher order statistical dependencies between the speech and video.  To prepare the speech features for being combined with video features, we use a network block, called A-subnet, to extract and organize speech features in a 2D form.
We do not directly use the MFCC coefficients to generate the 2-D feature maps.  Instead, we adopt a three-layers bidirectional LSTM module to extract features from the MFCC coefficients.
That is $\mathcal{L}_{t \pm m} = \text{LSTM}(\mathcal{A}_{t \pm m})$, where $\mathcal{L}_{t \pm m}=\{L_{t-m},...,L_{t+m}\}$ is the extracted LSTM feature sequence of equal length to $\mathcal{A}_{t \pm m}$.

The A-subnet takes the $\mathcal{L}_{t \pm m}$ as input and outputs a cluster of 2-D feature maps of the same size as the video frame.  The sub-network consists of one linear layer and five upsampling residual blocks, are illustrated in Fig.~\ref{subnets}.


\subsection{Video feature extraction}
In order to fully exploit spatiotemporal correlations in video signals, our network takes a group of consecutive $(2n+1)$ compressed frames $\{Y_{t-n},...,Y_{t+n}\}$ as the second input in addition to associated speech.  Due to motions of camera or/and object (head in our case), the current frame $Y_t$ and its neighboring frames are misaligned.  Aligning these video frames helps the CNN blocks for feature extraction to learn or predict spatial details more accurately.  In recent studies on video super-resolution, Tian~\textit{et~al.}~\cite{TDAN} and Wang~\textit{et~al.}~\cite{EDVR} proposed to use deformable convolution~\cite{dcn} to align each neighboring frame to the reference frame and have achieved the state-of-the-art performance in video super-resolution task.
Inspired by their success, we also adopted the deformable convolution to align the current frame and its neighboring frames in the MDVD-Net.

However, in case of severe degradation of video quality, the alignment of neighboring frames is still inaccurate even after introducing deformable convolution.  To improve the alignment accuracy, we use motion vectors (MVs) in compression code stream as clues to guide the deformable convolution.  Motion vectors, though noisy and block-level represented, constitute a rough approximation of optical flows, hence they are valuable information on inter-frame dependencies.  Another reason of using motion vectors is operational. Unlike dense optical flows that require extensive computations, motion vectors are readily available from the code stream, free of computational cost.


\begin{figure}[t]
  \centering
  \includegraphics[width=\linewidth]{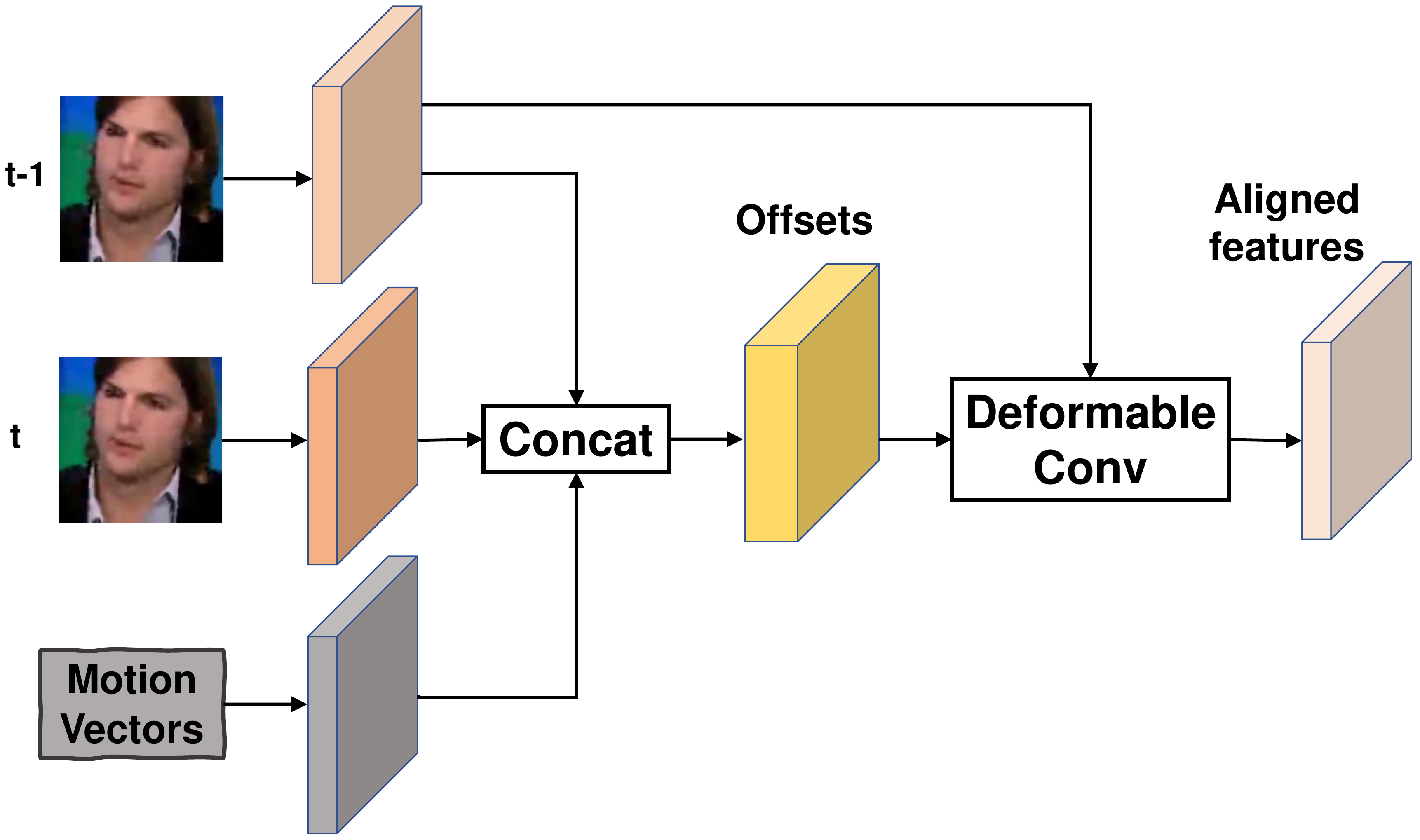}
  \vspace{-0.5cm}
  \caption{MV-guided alignment with deformable convolutions.}
  \label{alignment}
\end{figure}

The MV-guided alignment with deformable convolution is explained in Fig.~\ref{alignment}.
Deformable alignment is performed in the CNN feature space not pixel domain.
Different from standard convolution which adopts the regular sampling grid, deformable convolution augments the sampling grid with learnable offsets $\Delta P$.
In our design, the learnable offsets $\Delta P$ are predicted from the concatenated features of neighboring frames and the corresponding motion vectors by several convolution layers, that is:
\begin{align}
\Delta P = Conv([F_t, F_{t-1}, MV])
\end{align}
where $F_t$ and $F_{t-1}$ are features of frame $Y_t$ and $Y_{t-1}$, respectively.
Then we apply the predicted offsets $\Delta P$ in the deformable convolutions to get the aligned features:
\begin{align}
  F_{t-1}^g = DConv(F_{t-1}, \Delta P)
\end{align}
where $DConv$ is the deformable convolution operator, $F_{t-1}^g$ is the aligned features of frame $Y_{t-1}$.
After alignment, a network block called V-subnet is designated to extract features from the aligned video frames. The V-subnet is detailed in Fig.~\ref{subnets}.

\begin{figure}[t]
  \centering
  \includegraphics[width=\linewidth]{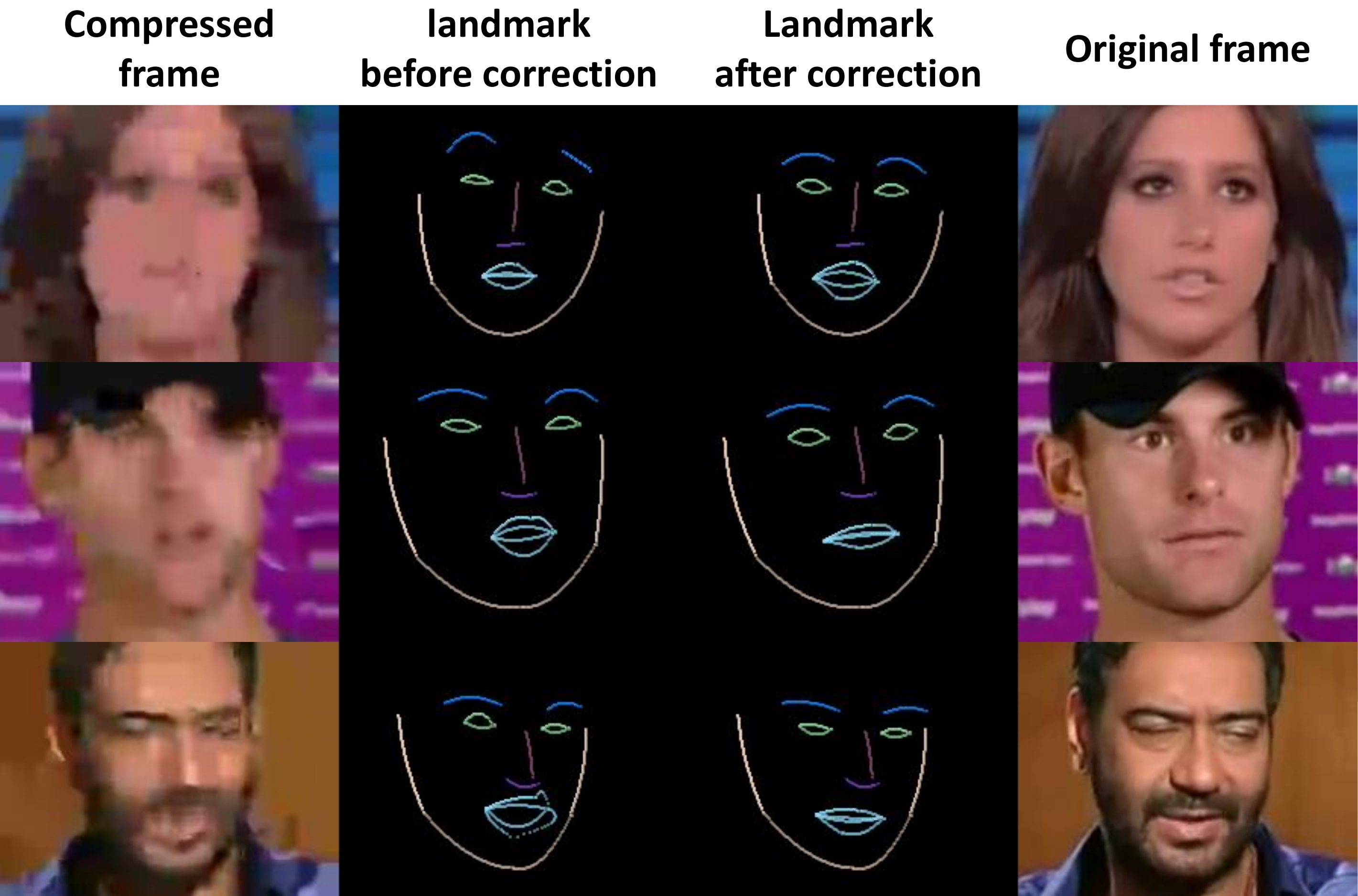}
  \caption{Facial landmarks before and after correction by speech.}
  \label{landmarks}
\end{figure}

\subsection{Landmark feature extraction}
In addition to directly learning 2D features from speech, we use facial landmark points as intermediate representations and associate them with speech signals. The degree of freedoms for facial landmarks is in the order of tens (68 in our implementation), as opposed to millions of pixels in conventional 2D feature maps.
This makes our method more robust even if trained with small datasets.
However, facial landmarks detected from aggressively compressed videos are very noisy and hence cannot be directly used to guide the face restoration.
In this case, the accompanying speech signal can be used to denoise the facial landmarks, especially in the region of mouth and eyes.  Moreover, the images of talking head video only not depend on linguistic contents, but also on the emotion of the speaker. The facial expression or the emotion of the speaker puts structural constraints on the shapes and relative positioning of eyes and eyebrows in addition to mouth.  Hence, the speech can shape not only the mouth region but also the eye region of the face image.
Specifically, we adopt LSTM to polish the landmarks as it is suited to model temporal dependencies between the utterance and moving facial landmarks.
Specifically, at each time instance $t$, the LSTM module takes as input the speech content within a window  $\{A_{t-m},...,A_{t+m}\}$; the output from LSTM layers is fed into a Multi-Layer Perceptron (MLP) and finally predicts the landmark displacements $\Delta L_t$.  These displacements are used to improve the initially estimated landmarks $L_t$ to $\hat{L_t} = L_t + \Delta L_t$.

Fig.~\ref{landmarks} demonstrates the effectiveness of speech signal in refining facial landmarks, by comparing the results before and after speech-guided correction.  The corrected facial landmarks are fed to the L-subnet (see Fig.~\ref{subnets}(c)) to extract features for being combined with those of other modalities.

\subsection{Spatial attention fusion}
After extracting time synchronized video and speech features, the next task is to fuse them for the purpose of removing compression artifacts. A simple approach is
to concatenate the video and speech features directly. However, the speech signal correlates most strongly to the image parts around the
mouth (e.g., lips, cheeks, and chin), instead of the entire face.  Therefore, the 2-D feature maps generated from the speech signal are not equally important in the spatial domain; they should be judiciously used to guide the video reconstruction of the mouth region.

\begin{figure}[t]
  \centering
  \includegraphics[width=\columnwidth, height=0.6\columnwidth]{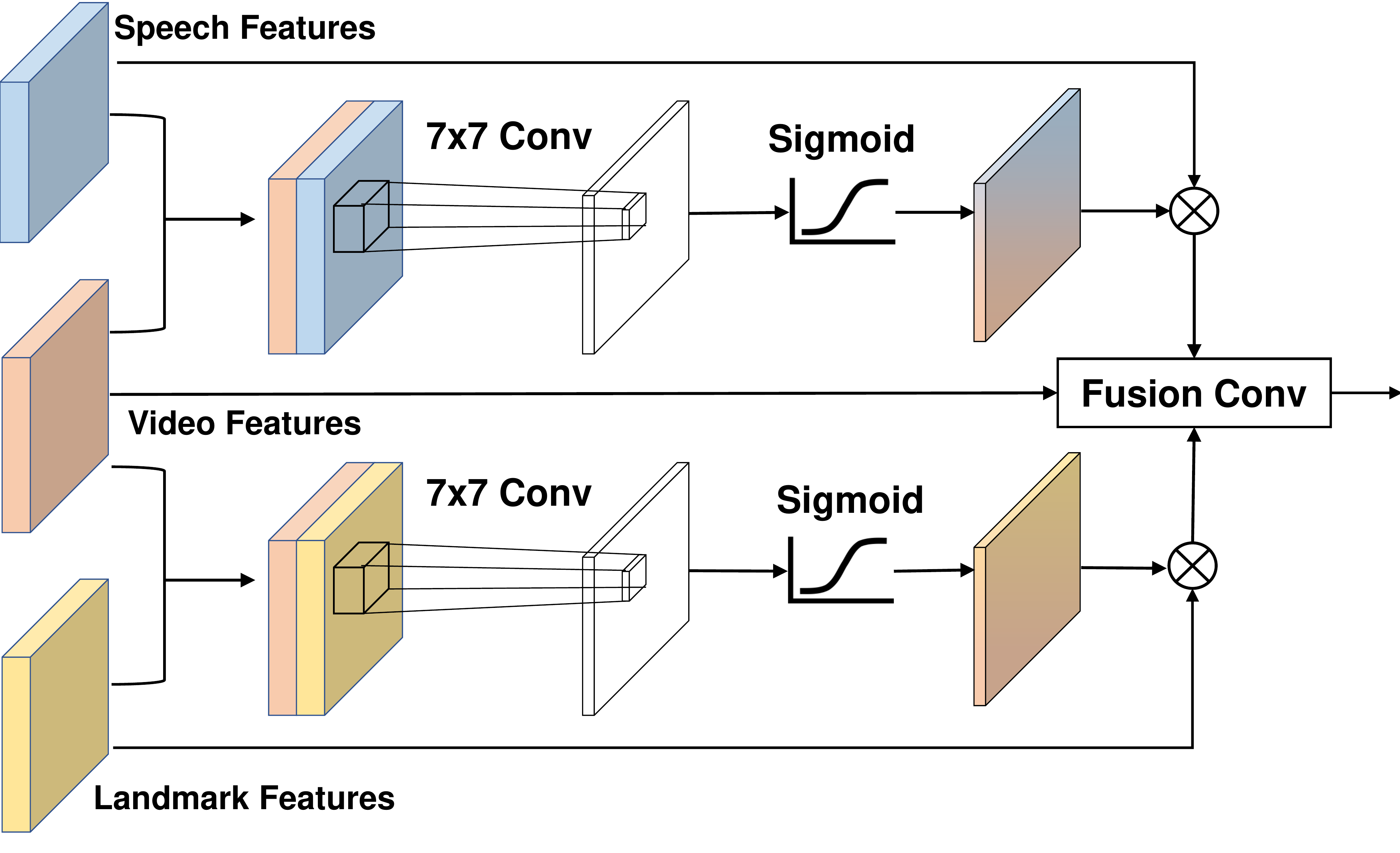}
  \caption{The architecture of the spatial attention fusion module.}
  \label{attention}
\end{figure}

However, natural head movements during talking change the position and even orientation of the speaker's mouth.  To capture such dynamics the network needs to temporally adjust speech feature maps and video feature maps at pixel level.
In addition, facial landmark features also need to be fused with the speech and video features in a temporally adaptive manner.
To this end, we introduce a spatial attention fusion module to allow time varying associations of speech, video and landmark features, as illustrated in Fig.~\ref{attention}.

In our design, the network computes an attention map from the speech and video features with a range from 0 to 1, where 0 represents that the speech signal feature at this position is completely useless for recovery and 1 means most critical. Same goes for the facial landmark features.
The spatial attention maps $M_t^a$ and $M_t^l$ are formulated as:
\begin{equation}
\begin{aligned}
M_t^a = \sigma(f^{7\times7}(\ [F_t^v, F_t^a]\ )\ ) \\
M_t^l = \sigma(f^{7\times7}(\ [F_t^v, F_t^l]\ )\ )
\end{aligned}
\end{equation}
where $F_t^v$, $F_t^a$ and $F_t^l$ are the feature maps generated from video, speech and landmarks, respectively.
$f^{7\times7}$ represents a convolution operation with the filter size of $7\times7$.
$\sigma$ denotes the sigmoid function, which is used to restrict the outputs $M_t^a$ and $M_t^l$ in [0, 1].  The $[\cdot,\cdot]$ denotes the concatenation operation.

The speech feature maps $F_t^a$ and landmark feature maps $F_t^l$ are then multiplied in a pixel-wise manner by the corresponding spatial attention maps, and then aggregated with the video feature maps $F_t^v$ using a few convolutional layers, that is:
\begin{align}
\hat{F}_t^a = F_t^a \odot M_t^a
\end{align}
\vspace{-0.7cm}
\begin{align}
\hat{F}_t^l = F_t^l \odot M_t^l
\end{align}
\vspace{-0.7cm}
\begin{align}
F_{agg} = Conv(\ [F_t^v, \hat{F}_t^a, \hat{F}_t^l]\ )
\end{align}
where $\hat{F}_t^a$, $\hat{F}_t^l$ are the attention-modulated feature maps, and $F_{agg}$ is the aggregated feature maps from the three modalities. Then as illustrated in Fig.~\ref{overall}, we feed the aggregated feature maps $F_{agg}$ into a reconstruction module, which is a UNet-like Encoder-Decoder network.

\subsection{Codec information branch}
Most existing methods for reducing video compression artifacts operate on the decoded frames only, ignoring the prior codec information available in the code stream.  Some
researchers~\cite{partition,DKFN,prior,lin2019partition} realized the benefit of the prior codec information to deep video decompression.  But they only fed the encoding prior like prediction residuals into neural networks along with the decoded frames, which is straightforward to do but has limited effect.  A potentially highly profitable piece of prior information is left unexploited: the DCT coefficient quantization intervals, which can be extracted from the compression code stream.
These quantizer structural data can be used to reduce the uncertainty of the latent video to the decoder. Specifically, we add a back projection module to bound the solution space by quantization intervals in the DCT transform domain, and thus refine the reconstructed video.
This back projection module is implemented using a piece-wise linear activation function embedded in the neural network.

In modern video compression standards, prediction based coding is a core operation.  Given an original frame $X_t$ to be coded, inter/intra frame prediction techniques are used to obtain a prediction frame of $X_t$, denoted by $P_t$. Then the prediction residual $E_t=X_t-P_t$ will be transformed into the DCT domain and quantized, followed by the entropy coding.
In the encoding phase, the DCT coefficients of $E_t$ (denoted by $E_t^{dct}$) are divided by a quantization table $Q$, and are then rounded to the nearest integers. When decoding, the decoder performs decompression by multiplying back the quantization table $Q$ in the DCT domain. The closed quantization and dequantization loop can be formulated as:
\begin{align}
\label{quan}
\centering
\hat{E}_t^{dct} =\left[ (E_t^{dct}) / Q \right] * Q
\end{align}
where $[\cdot]$ represents the round operation and $\hat{E}_t^{dct}$ denotes the decoded DCT coefficients of the prediction residual block.
The decoded frame is obtained by inverse DCT transform of $\hat{E}_t^{dct}$ and adding the results to the prediction frame, that is $Y_t=P_t+\hat{E_t}$, where $\hat{E_t} = \text{IDCT}(\hat{E}_t^{dct})$.

Eq.\ref{quan} provides the following DCT coefficient range constraint:
\begin{align}
\hat{E}_t^{dct} - Q/2 \leq E_t^{dct} \leq \hat{E}_t^{dct} + Q/2
\end{align}
That is, from the decoded DCT coefficients of the prediction errors, we can derive the lower and upper bounds of the original DCT coefficients of the prediction residuals.

\begin{figure}[t]
  \centering
  \includegraphics[width=\columnwidth]{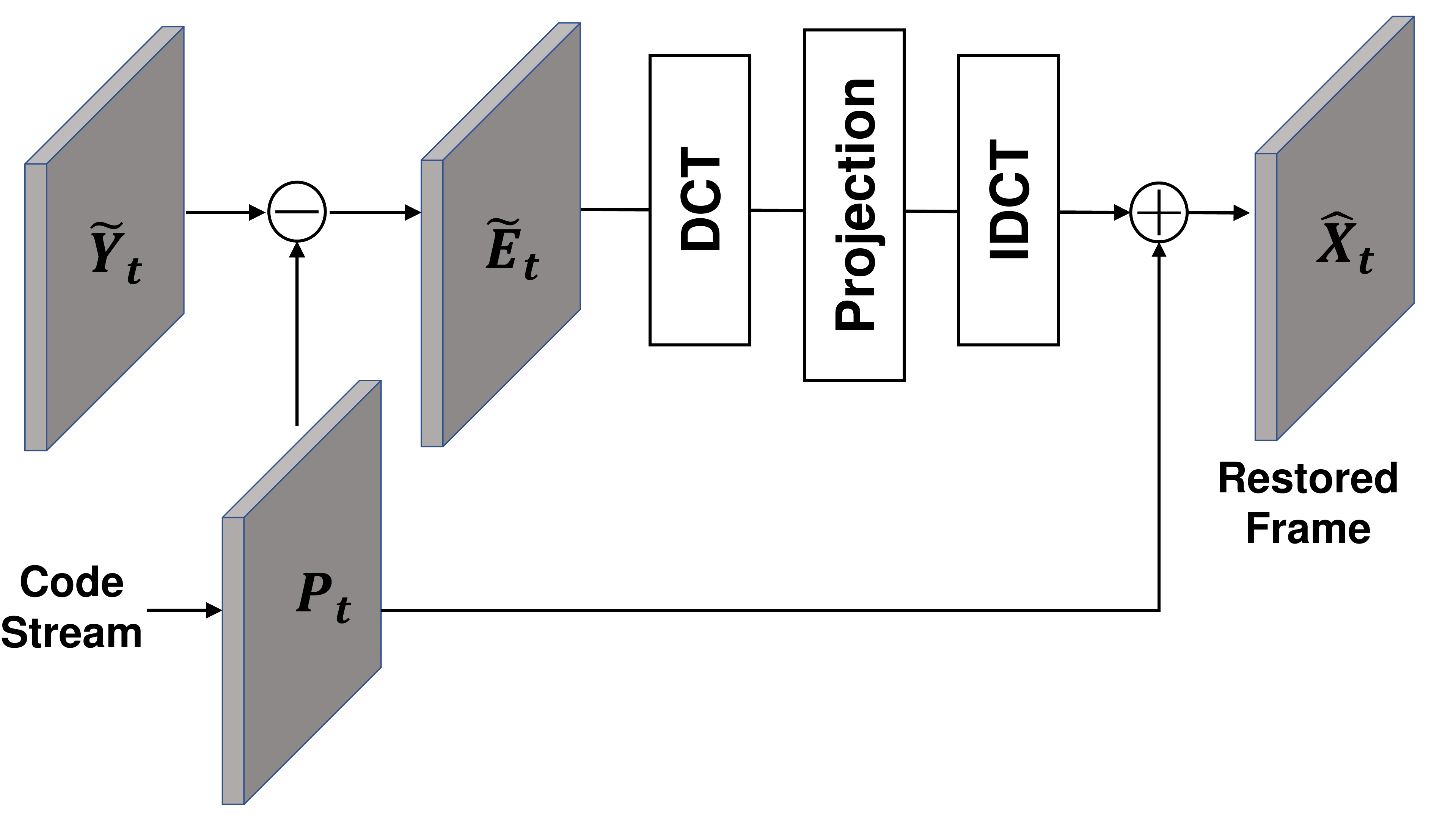}
  \caption{The architecture of the back projection module.}
  \label{projection}
\end{figure}

We can enforce the DCT coefficient bounds in the decision of the MDVD-Net by inserting a back projection module before the final output (see Fig~\ref{projection}).
Letting $\tilde{Y}_t$ be the output of the reconstruction module, the back projection module imposes constrains on the DCT coefficients of $\tilde{E}_t = \tilde{Y}_t - P_t$:
\begin{align}
\centering
F(\tilde{E}_t^{dct}) =
\begin{cases}
L(i,j),                  & \tilde{E}_t^{dct}(i,j) < L(i,j) \\
\tilde{E}_t^{dct}(i,j),  & \tilde{E}_t^{dct} \in [L(i,j), U(i,j)] \\
U(i,j),                  & \tilde{E}_t^{dct}(i,j) > U(i,j)
\end{cases}
\end{align}
where $L=\hat{E}_t^{dct}-Q/2$ and $U=\hat{E}_t^{dct}+Q/2$, $i$ and $j$ are the quantizer indexes in the DCT domain.  The back projection function $F(\cdot)$ can be implemented as a piecewise linear activation in the neural network.  Finally, the reconstructed frame $\hat{X}_t$ is given by
\begin{align}
\centering
\hat{X}_t = \text{IDCT}(F(\tilde{E}_t^{dct}))+P_t.
\end{align}

\section{Experiments}
To systematically evaluate and analyze the performance of the proposed MDVD-Net methodology conditioned on priors of different strengths, we conduct extensive experiments on two datasets: (1) Obama dataset~\cite{obama} containing single person; (2) VoxCeleb2 dataset~\cite{vox2} which contains multiple persons.  In these two sets of experiments, the MDVD-Net is trained for compressed videos of a particular known speaker and for generic talking heads, respectively.

\subsection{Data preparation}
\textbf{Obama Dataset.}
We collect 198 high-quality Barack Obama’s Weekly Address videos from YouTube. Each video is approximately three to six minutes long and 790 minutes in total.
This dataset is divided into two parts: 160 videos for training/validation, and the rest 38 videos for testing.
We detect and crop the face region from each frame and then resize it to $256\times256$ resolution.

\textbf{VoxCeleb2 Dataset.}
VoxCeleb2~\cite{voxceleb, voxceleb2} is an audio-visual dataset consisting of short clips of human speech, extracted from interview videos uploaded to YouTube. It contains speech from speakers spanning a wide range of different ethnicities, accents, professions and ages. All speaking face-tracks are captured "in the wild", with background chatter, laughter, overlapping speech, pose variation and different lighting conditions.
Specifically, VoxCeleb2 contains over 1 million utterances for 6,112 celebrities.
we use the VoxCeleb2 development set for training and evaluate the trained model on the VoxCeleb2 test set.


The compressed videos are all generated by FFmpeg with h264 and h265 video codecs in constant bit rate (CBR) mode. We set bit rate at different levels (60kbps, 70kbps, 80kbps, 90kbps, 100kbps, 110kbps, 120kbps) for a comprehensive evaluation.

\subsection{Training details}
We carry out an end-to-end training of all modules presented in the MDVD-Net, except for the
landmark correction module and back projection module, in which the former is pretrained using extra data and the latter has no parameters to be learned.
All cropped face images are resized to $256\times256$ for training and testing.
The window size for video signal is 5 and for speech signal is 21. That is,
$\mathcal{Y}_{t \pm 2} = \{Y_{t-2}, Y_{t-1}, Y_{t}, Y_{t+1}, Y_{t+2}\}$ and
$\mathcal{A}_{t \pm 10} = \{A_{t-10}, A_{t-9}, ..., A_{t+9}, A_{t+10}\}$.
The training loss is set to $L_1$ loss, defined by $L_1(X_t, \hat{X}_t)=||X_t - \hat{X}_t||_1$.

We train the MDVD-Net with Adam optimizer~\cite{adam} by setting $\beta_1=0.9$ and $\beta_2=0.999$, with initializing learning rate as $1\times10^{-4}$. Mini-batch size is set to 32.
We implement the proposed MDVD-Net model in PyTorch~\cite{pytorch} and train it with four NVIDIA 2080 Ti GPUs. The training takes about one week to converge on the VoxCeleb2 dataset.

\subsection{Extraction of video codec information}
\begin{figure}[t]
  \centering
  \includegraphics[width=\columnwidth]{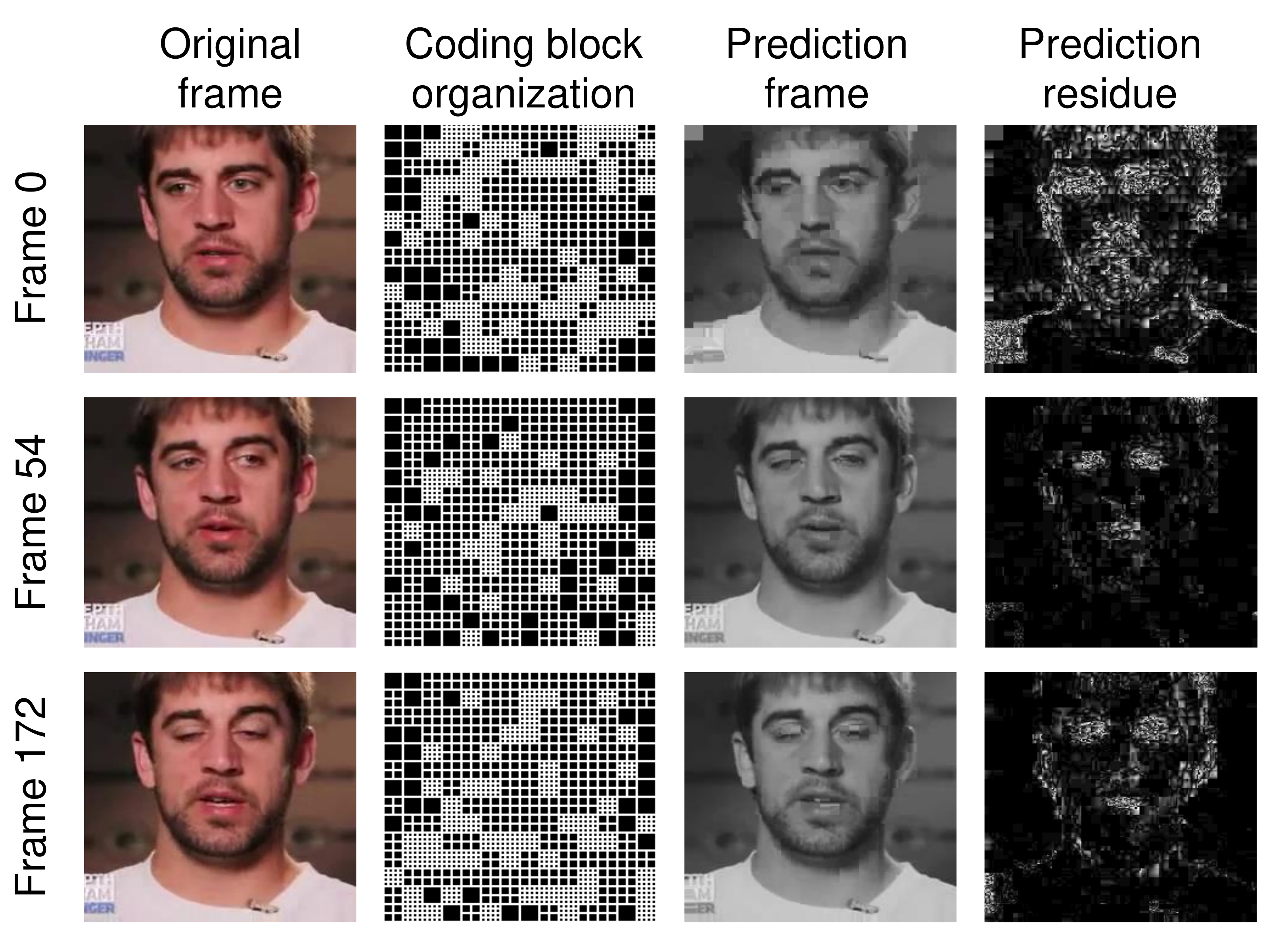}
  \caption{The illustration of codec information of the H.264 video compression standard.
  From left to right are: original frame, transform unit (TU) partition, prediction frame in Y channel, and prediction residue in Y channel, respectively.}
  \label{codec_demo}
\end{figure}

In order to implement the proposed back projection module, codec structural information like the DCT transform block partition, prediction frame and prediction residual image are required.
In video compression standards, the pixels are organized by a hierarchical block structure, and the transform block is the basic coding unit.
However, the common video decoding tools like FFmpeg is not able to extract the DCT transform block partition and other encoding information from the code stream.  To overcome this difficulty and get all codec priors required by the projection module, we developed a tool to extract the encoding information from the compressed code stream, including transform block partition, prediction frame and prediction residual, etc.  Some pieces of encoding information of H.264 (prediction frame, prediction residual and transform unit (TU) partition are shown in~Fig.\ref{codec_demo}.

\begin{figure*}[t]
  \centering
  \begin{subfigure}[b]{0.49\textwidth}
    \centering
    \includegraphics[width=\textwidth]{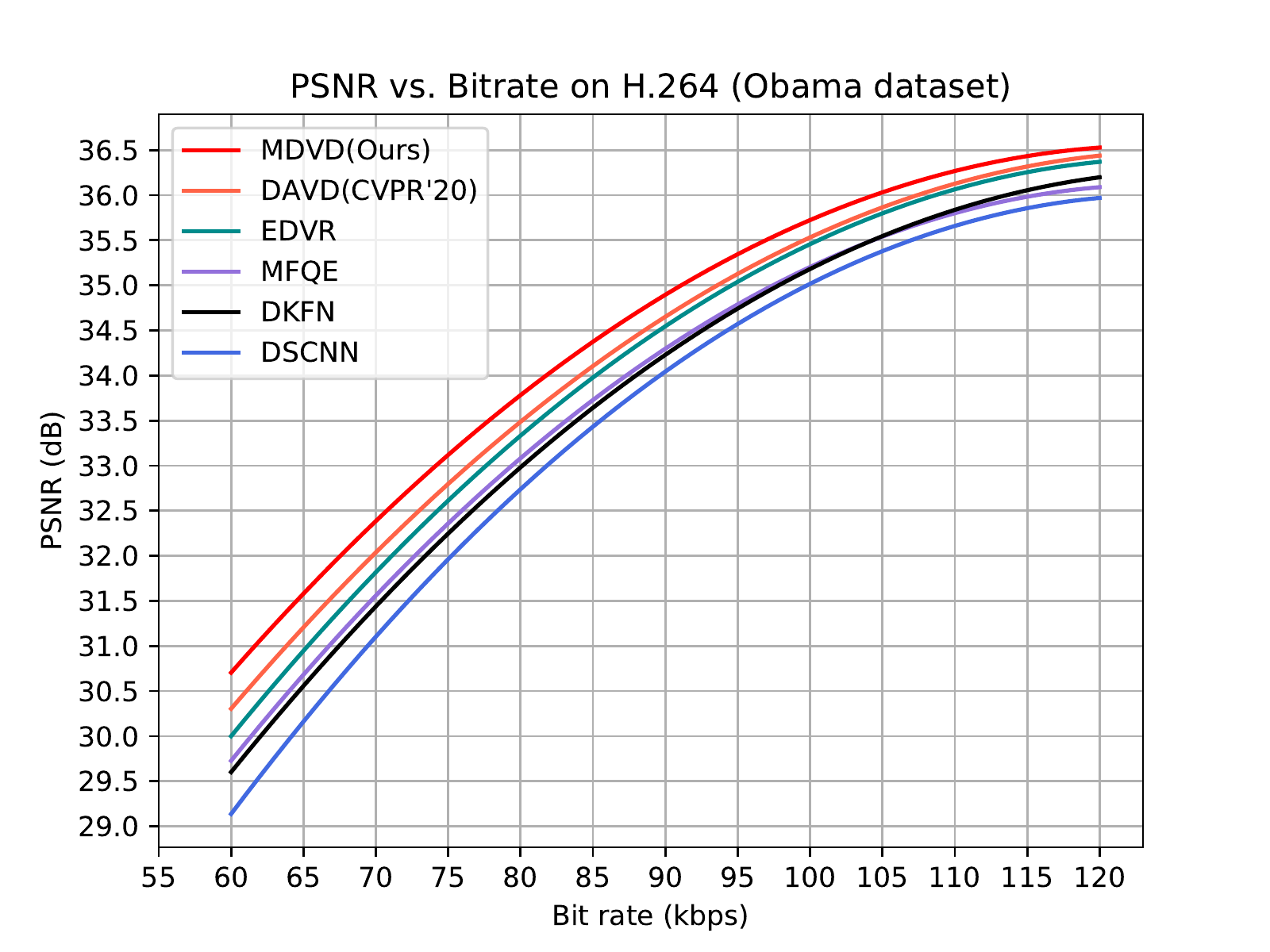}
    \caption{}
    \label{rd_obama_psnr}
  \end{subfigure}
  \hfill
  \begin{subfigure}[b]{0.49\textwidth}
    \centering
    \includegraphics[width=\textwidth]{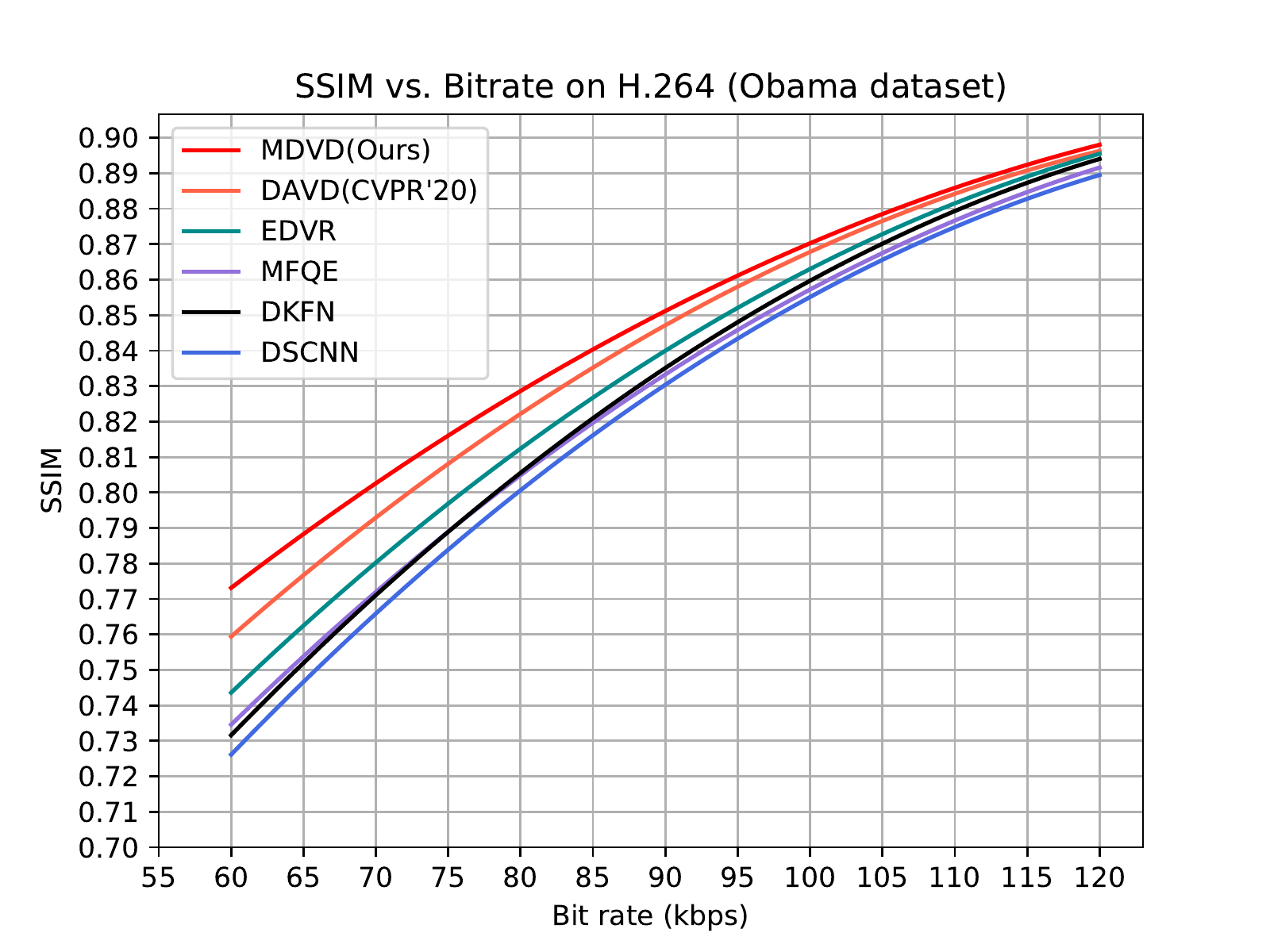}
    \caption{}
    \label{rd_obama_ssim}
  \end{subfigure}
  \\
  \begin{subfigure}[b]{0.49\textwidth}
    \centering
    \includegraphics[width=\textwidth]{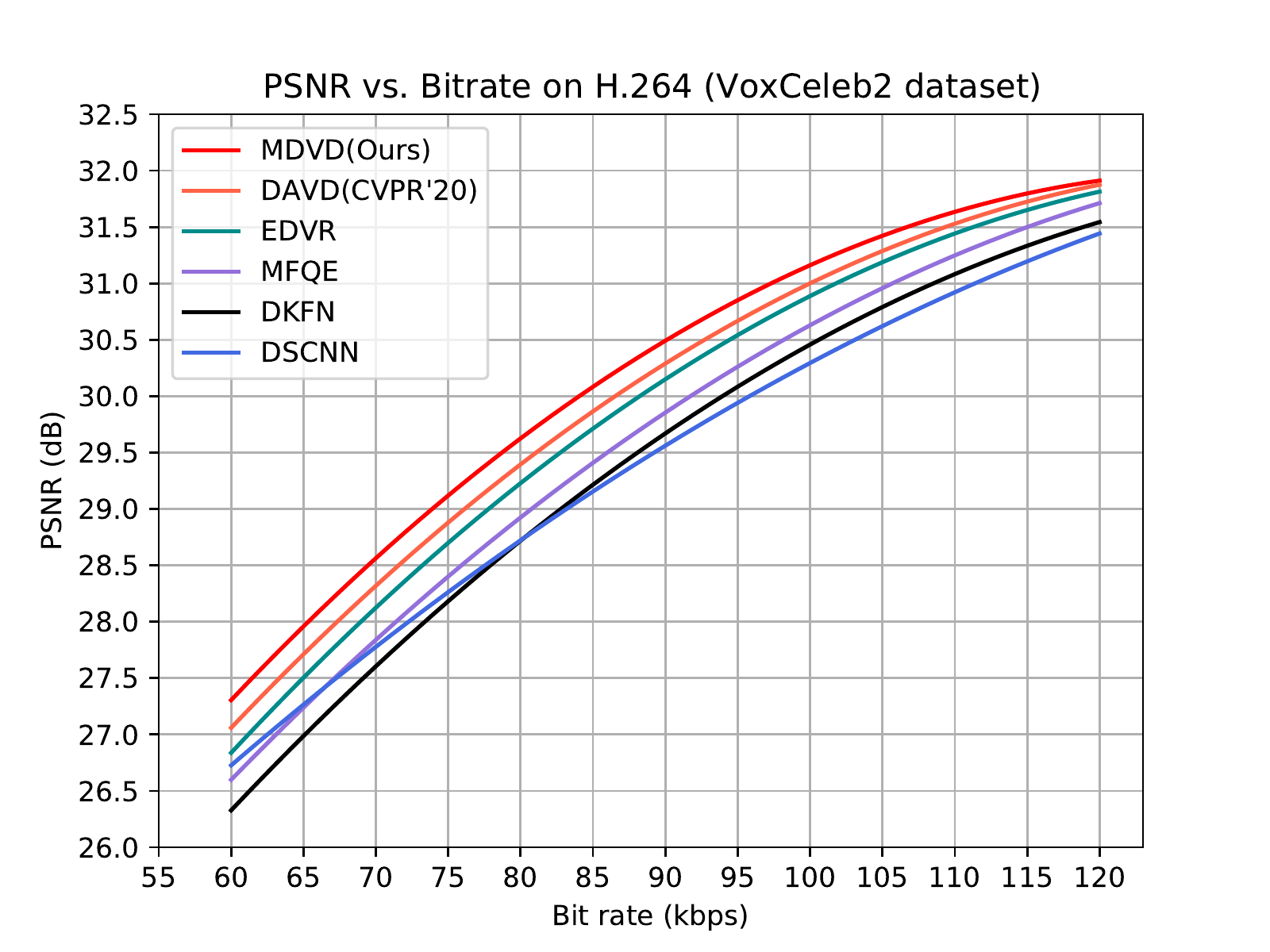}
    \caption{}
    \label{rd_vox2_psnr}
  \end{subfigure}
  \hfill
  \begin{subfigure}[b]{0.49\textwidth}
    \centering
    \includegraphics[width=\textwidth]{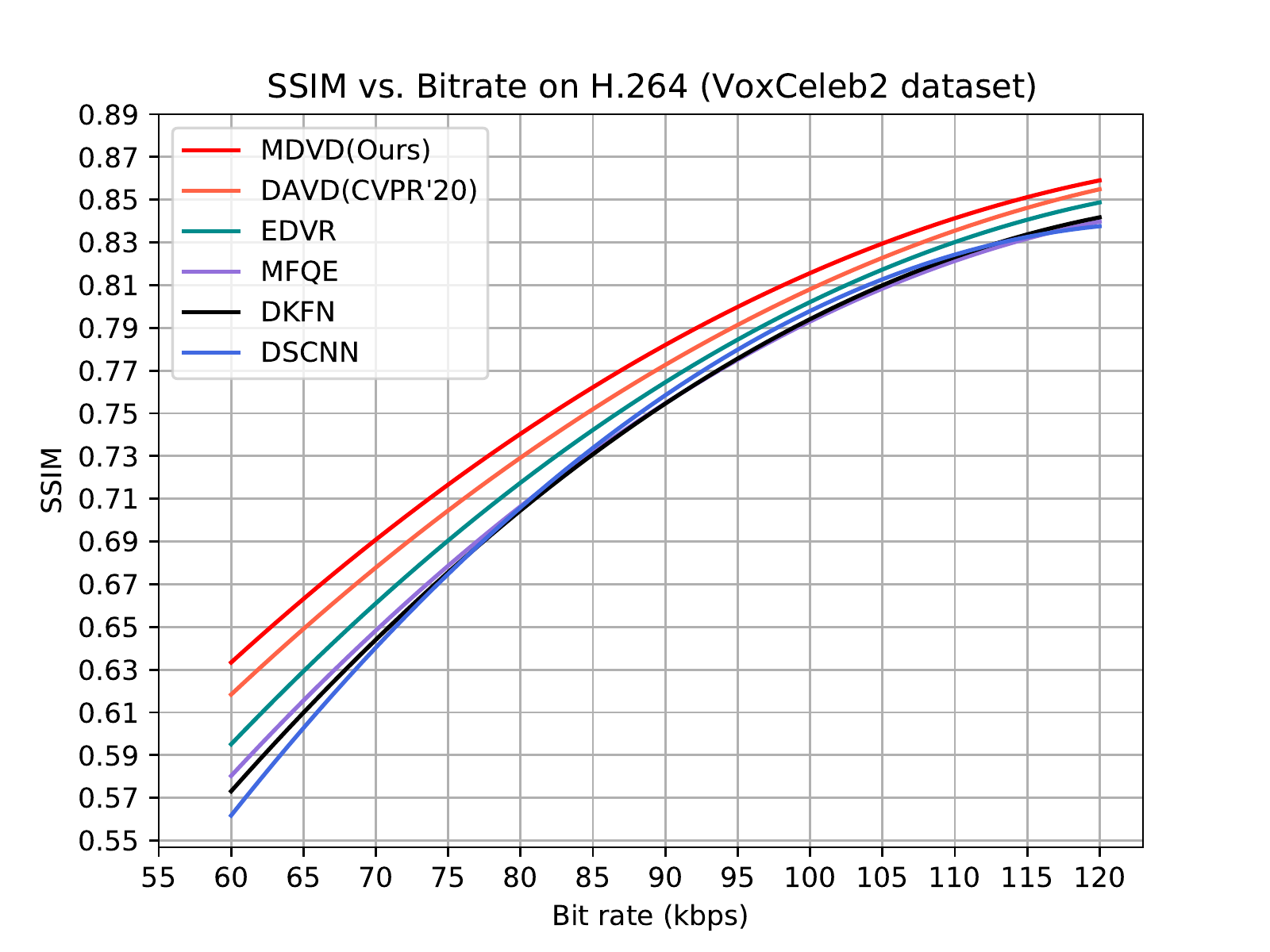}
    \caption{}
    \label{rd_vox2_ssim}
  \end{subfigure}
  \caption{Rate-distortion curves of the competing methods on the Obama and VoxCeleb2 dataset on H.264 video codec. The proposed MDVD-Net cleayly outperforms all existing methods by a large margin.}
  \label{rds_264}
\end{figure*}

\begin{figure*}[t]
  \centering
  \begin{subfigure}[b]{0.49\textwidth}
    \centering
    \includegraphics[width=\textwidth]{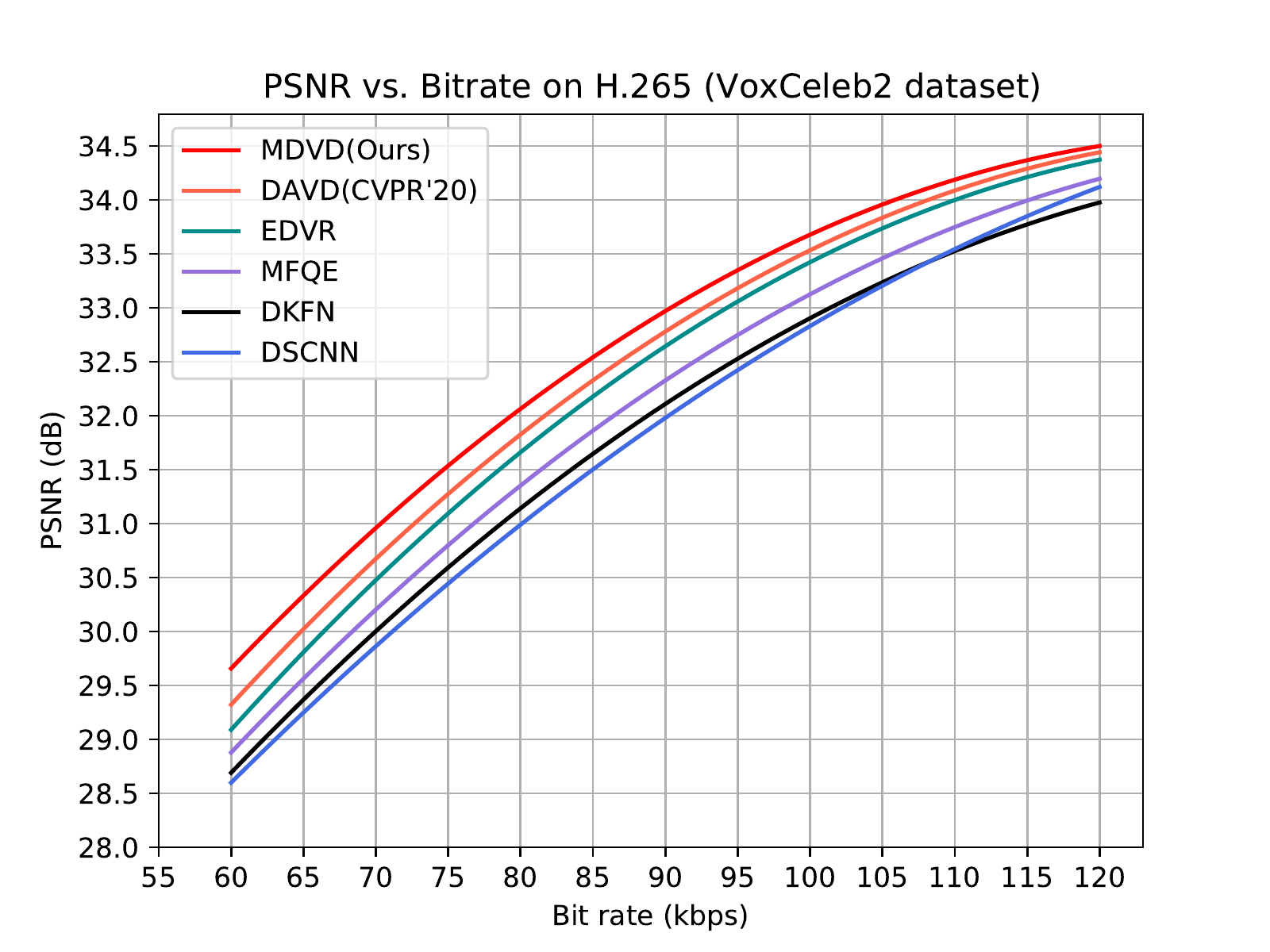}
    \caption{}
    \label{rd_vox2_psnr_265}
  \end{subfigure}
  \hfill
  \begin{subfigure}[b]{0.49\textwidth}
    \centering
    \includegraphics[width=\textwidth]{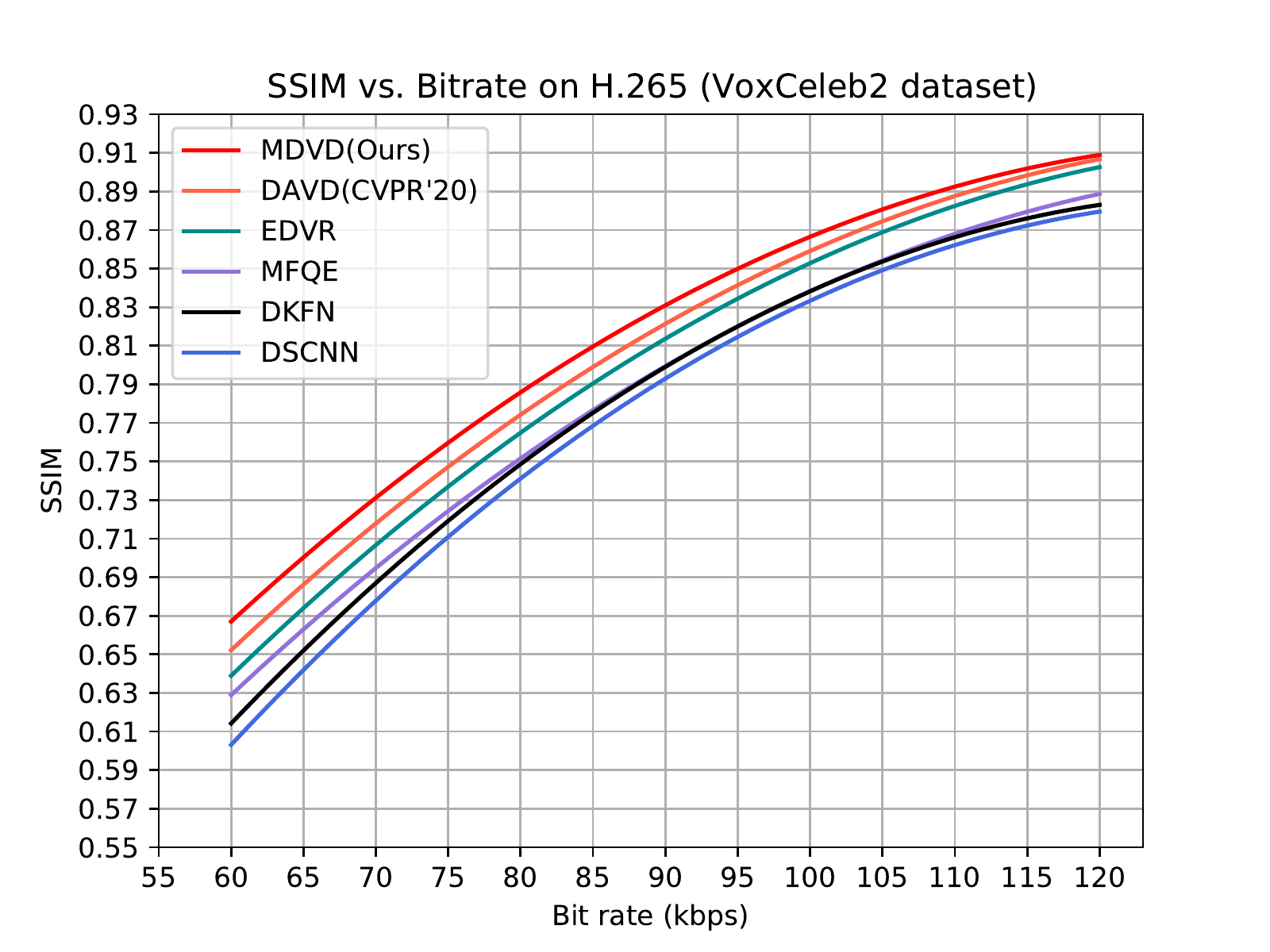}
    \caption{}
    \label{rd_vox2_ssim_265}
  \end{subfigure}
  \caption{Rate-distortion curves of the competing methods on the VoxCeleb2 dataset on H.265 video codec. The proposed MDVD-Net cleayly outperforms all existing methods by a large margin.}
  \label{rds_265}
\end{figure*}

\begin{figure*}[t]
  \centering
  \includegraphics[width=0.92\linewidth]{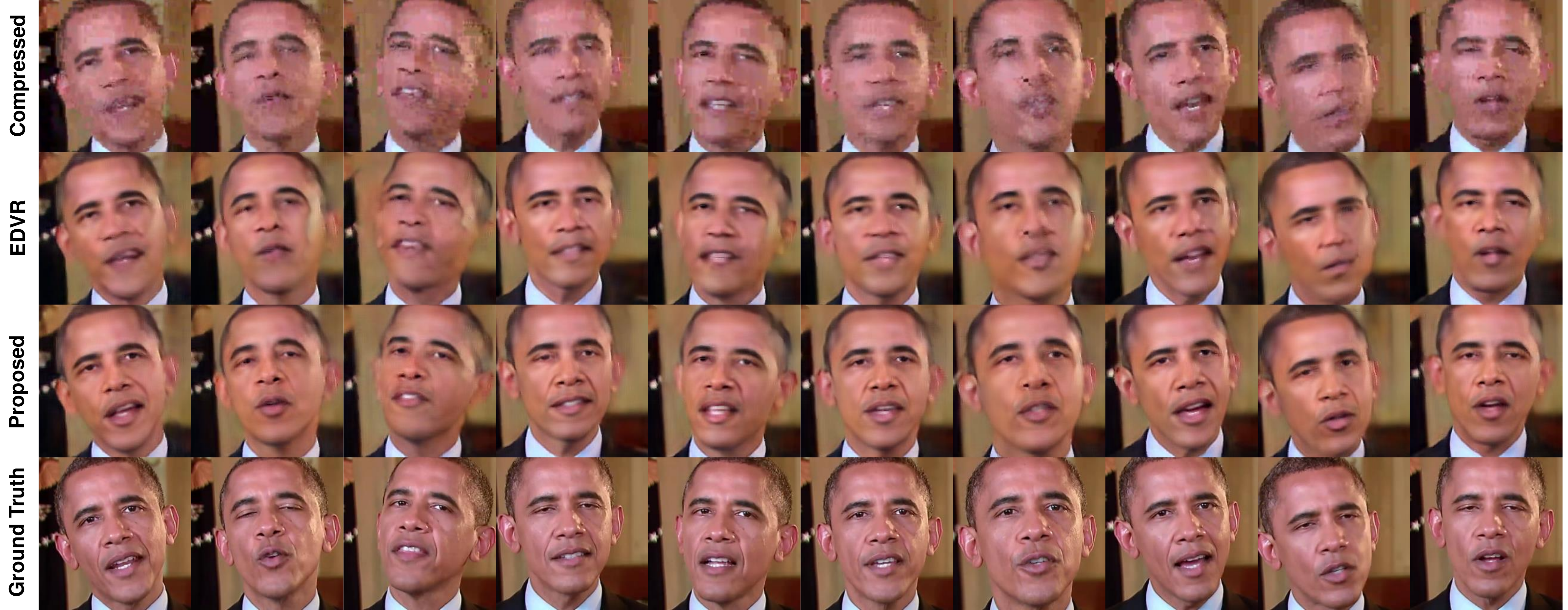}
  \caption{Visual comparisons of differnet methods on the Obama dataset.}
  \label{obama_eva_1}
\end{figure*}

\begin{figure*}[!h]
  \centering
  \begin{subfigure}[t]{\linewidth}
    \centering
    \includegraphics[width=0.92\linewidth]{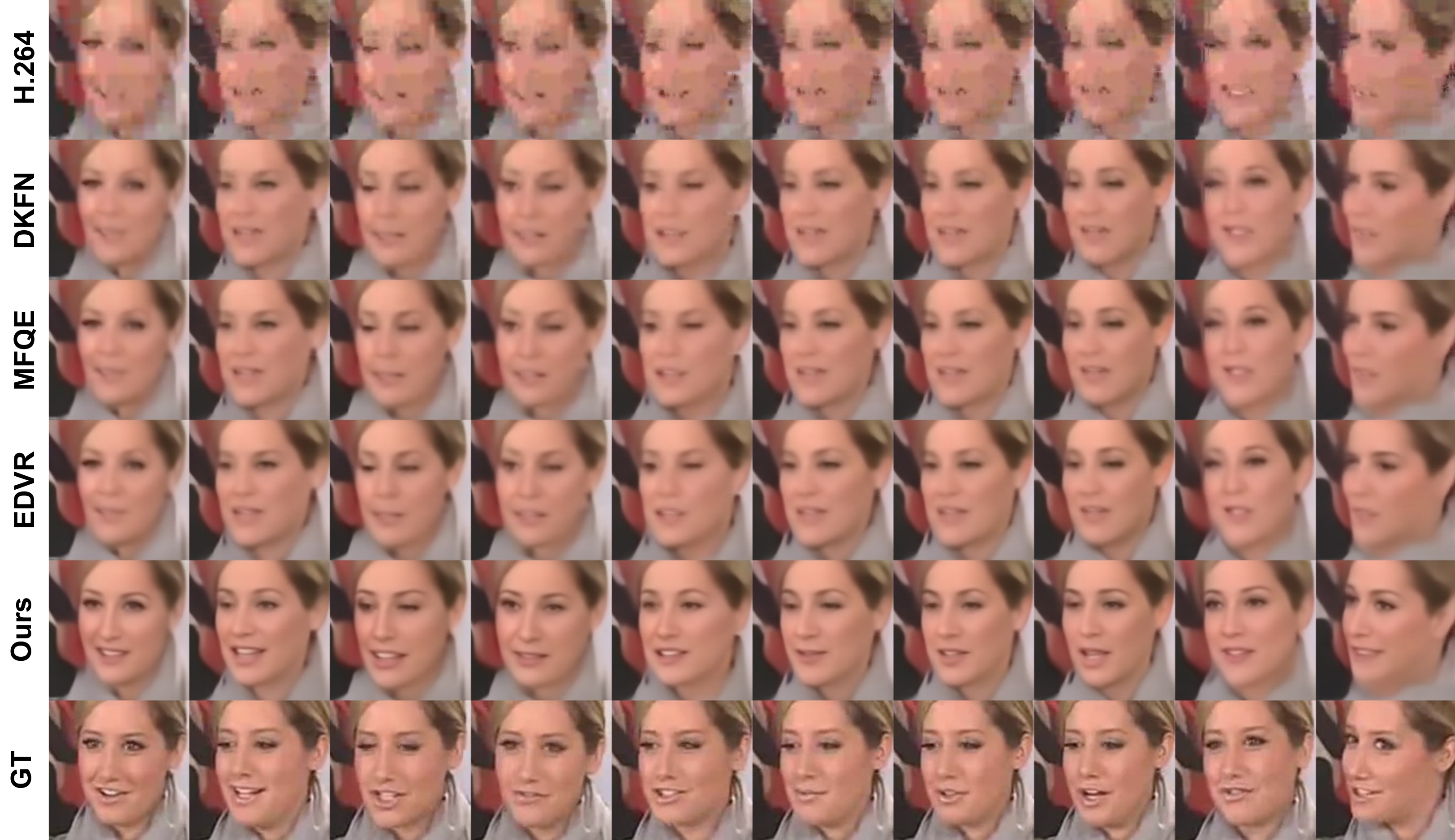}
    \caption{}
    \label{vox2_eva_1}
  \end{subfigure}
  \\
  \begin{subfigure}[t]{\linewidth}
    \centering
    \includegraphics[width=0.92\linewidth]{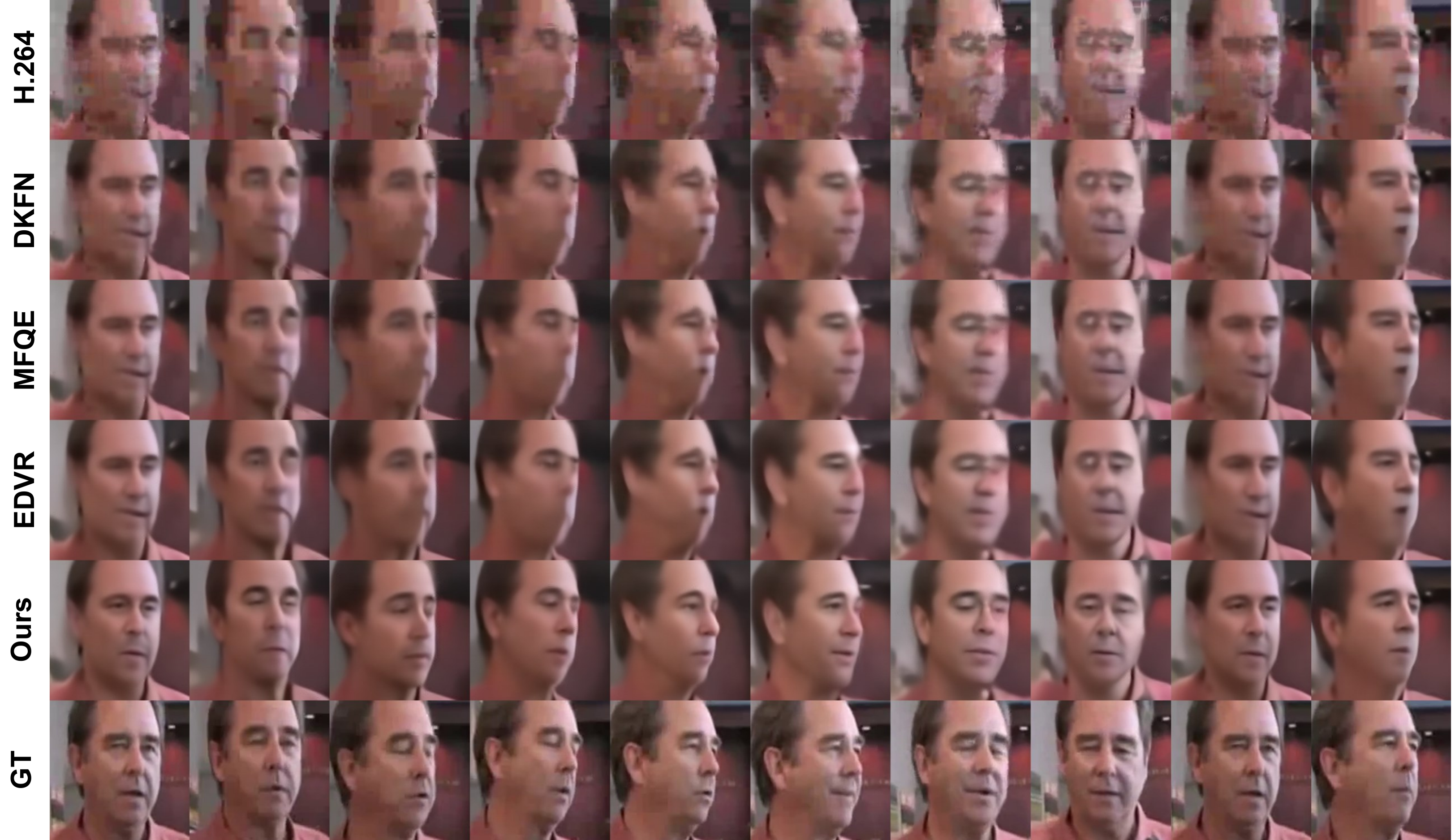}
    \caption{}
    \label{vox2_eva_2}
  \end{subfigure}
  \caption{Visual comparisons of differnet methods on the VoxCeleb2 dataset.}
  \label{vox2_eva}
\end{figure*}

\subsection{Comparison with state-of-the-art methods}
We compare our MDVD-Net with several state-of-the-art methods for video restoration:
DSCNN~\cite{dscnn}, DKFN~\cite{DKFN}, MFQE~\cite{MFQE} and EDVR~\cite{EDVR},
where DSCNN, DKFN and MFQE are particularly designed for the task of video compression artifacts removal and EDVR is claimed as a unified framework for generic video restoration tasks.

To demonstrate the advantages of tailoring the restoration network to talking heads over generic objects, we compare our network with the DKFN and MFQE models pretrained by the authors using generic videos (the only two pretrained models available to us).
On all three datasets, the MDVD-Net outperforms DKFN and MFQE by as much as 2.5dB.
This is not surprising, because the other two networks are all trained using generic video datasets like Vimeo-90K~\cite{TOFLOW} or JCT-VC~\cite{jctvc}.
In order to factor out the effects of different training sets, we retrain all CNN networks in the comparison group from scratch using the same datasets (Obama, VoxCeleb2 and Ravdess) in our experiments.

\textbf{Evaluation on Obama dataset.}
Fig.~\ref{rd_obama_psnr} and \ref{rd_obama_ssim} graphically presents the rate-distortion behaviours of different methods for the Obama dataset in PSNR and SSIM metrics. The proposed MDVD-Net cleayly outperforms all existing methods in terms of PSNR by a large margin. Notably, compared to the stat-of-the-art method EDVR, the gain achieved by MDVD-Net can reach 0.7dB when the bit rate is 60kbps, but will get smaller as the bit rate increases. When the bit rate increases to 120kbps, the performance gain drops to about 0.1dB. Compared to the conference version, the proposed MDVD beats DAVD in a large margin in both PSNR and SSIM.
This observation implies that the lower the bit rate, the greater the help of other modalities for face recovery.
We also provide the qualitiative comparisons with the stat-of-the-art method EDVR in Fig.~\ref{obama_eva_1}. It can be seen that the proposed MDVD-Net can restore facial features much better, such as more precise mouth shape, clearer teeth, sharper lips and muscle contours.
The above experimental results show that if the network is trained for a particular known speaker,
the assistance of other modalities (e.g., speech, landmark, codec  information) can bring significant performance gains on face video restoration task.

\textbf{Evaluation on VoxCeleb2 dataset.}
Fig.~\ref{rd_vox2_psnr}, \ref{rd_vox2_ssim} presents the qualitative results of competing methods achieved on the VoxCeleb2 dataset in PSNR and SSIM metrics. We also provide the rate-distortion curves on H.265 video codec in 
Fig.~\ref{rd_vox2_psnr_265} and \ref{rd_vox2_ssim_265} . Again, the proposed MDVD-Net surpasses all existing methods in terms of PSNR. But the improvement is slightly lesser than the case of training the network for a particular known speaker. The PSNR gain is about 0.4dB when the bit rate is 60kbps and falls to 0.1dB when the bit rate increases to 120kbps. Compared to the conference version, the newly proposed MDVD method achieves the reasonable performance gains
in both PSNR and SSIM.
The qualitative comparisons are shown in Fig.~\ref{vox2_eva}.  It can be seen that the proposed MDVD-Net achieves clearer face structure and richer details.  More experimental results can be found in the supplementary materials.  From these experiments one can see that after trained using large datasets of diverse talking head videos, the proposed MDVD-net can learn how to exploit the common correlations between face dynamics and the multimodality priors to improve the quality of decompressed videos.


\begin{table*}[t]
  \centering
  \caption{Quantitative results (PSNR) of ablation studies on the Obama dataset and VoxCeleb2 dataset.
  \textbf{VPB}: video processing branch;
  \textbf{MV}: motion vectors;
  \textbf{SPB}: speech processing branch;
  \textbf{FLWOC}: facial landmark without correction;
  \textbf{FLWC}: facial landmark with correction;
  \textbf{SAF}: spatial attention fusion;
  \textbf{VCI}: video codec informaion.}
  \label{ablation}
  \renewcommand\arraystretch{1.4}
  \begin{tabular}{ccccccccccccccccccc}
  \hline
  \multicolumn{8}{c}{Obama} & & \multicolumn{8}{c}{VoxCeleb2} \\
  \hline
  VPB & MV & SPB & FLWOC & FLWC & SAF & VCI & PSNR & & VPB & MV & SPB & FLWOC & FLWC & SAF & VCI & PSNR \\
  \hline
  \checkmark & & & & & & & 29.88 & & \checkmark & & & & & & & 26.82 \\
  \checkmark & \checkmark & & & & & & 29.96 & & \checkmark & \checkmark & & & & & & 26.85 \\
  \checkmark & \checkmark & \checkmark & & & & & 30.23 & & \checkmark & \checkmark & \checkmark & & & & & 27.01 \\
  \checkmark & \checkmark & \checkmark & \checkmark & & & & 30.31 & & \checkmark & \checkmark & \checkmark & \checkmark & & & & 27.04 \\
  \checkmark & \checkmark & \checkmark & \checkmark & & \checkmark & & 30.37 & & \checkmark & \checkmark & \checkmark & \checkmark & & \checkmark & & 27.06 \\
  \checkmark & \checkmark & \checkmark & & \checkmark & & & 30.43 & & \checkmark & \checkmark & \checkmark & & \checkmark & & & 27.12 \\
  \checkmark & \checkmark & \checkmark & & \checkmark & \checkmark & & 30.54 & & \checkmark & \checkmark & \checkmark & & \checkmark & \checkmark & & 27.20 \\
  \checkmark & \checkmark & \checkmark & & \checkmark &  \checkmark & \checkmark & 30.72 & & \checkmark & \checkmark & \checkmark & & \checkmark &  \checkmark & \checkmark & 27.31 \\
  \hline
  \end{tabular}
\end{table*}
\begin{figure*}[t]
  \centering
  \includegraphics[width=0.8\linewidth]{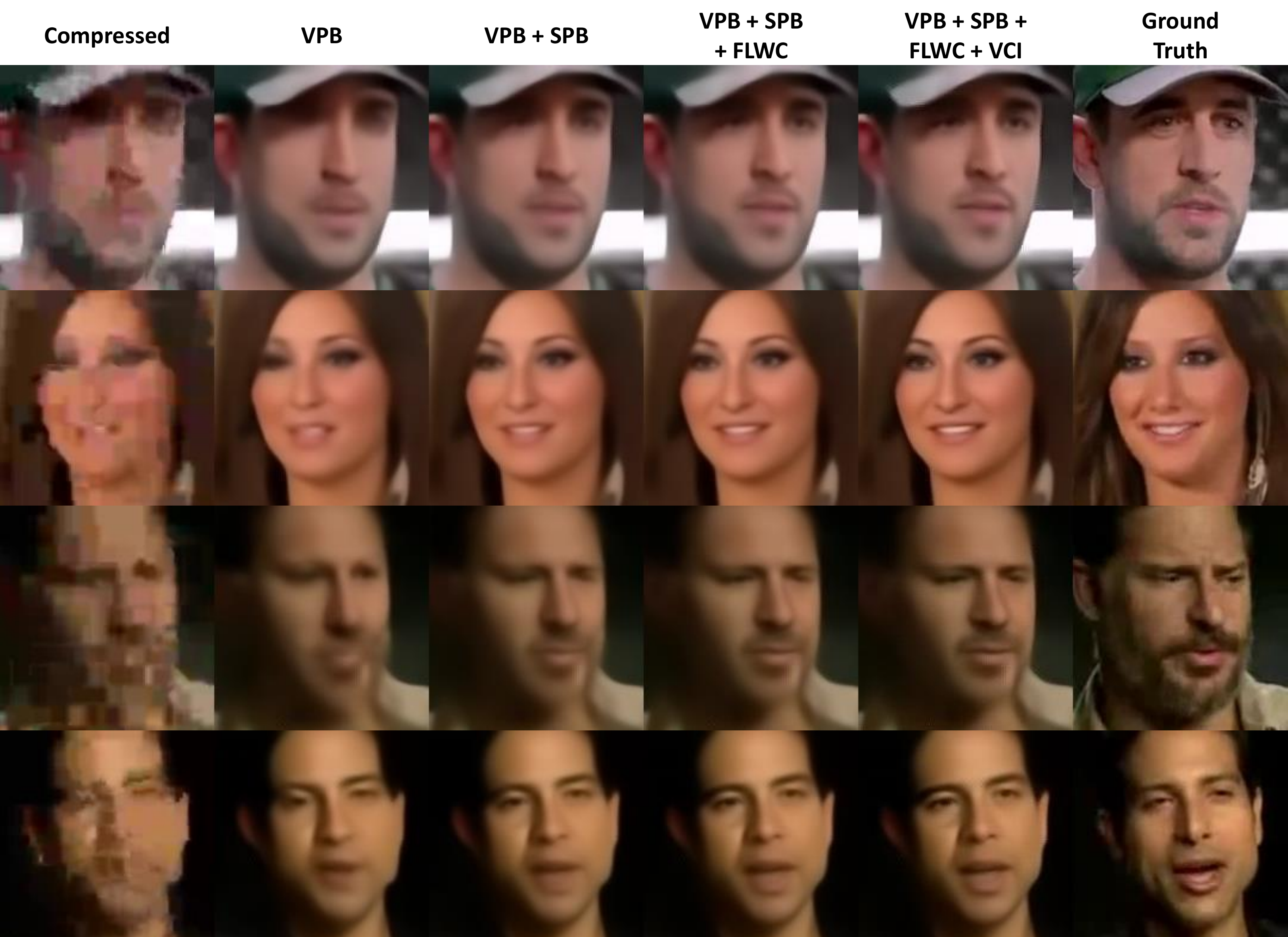}
  \caption{Visual comparisons of ablation studies on the VoxCeleb2 dataset. With the incorporating of various prior information from different modalities, the restoration result is getting better.
  \textbf{VPB}: video processing branch;
  \textbf{SPB}: speech processing branch;
  \textbf{FLWC}: facial landmark with correction;
  \textbf{VCI}: video codec informaion.}
  \label{abl_eva}
\end{figure*}
\begin{figure*}[!h]
  \centering
  \includegraphics[width=0.8\linewidth]{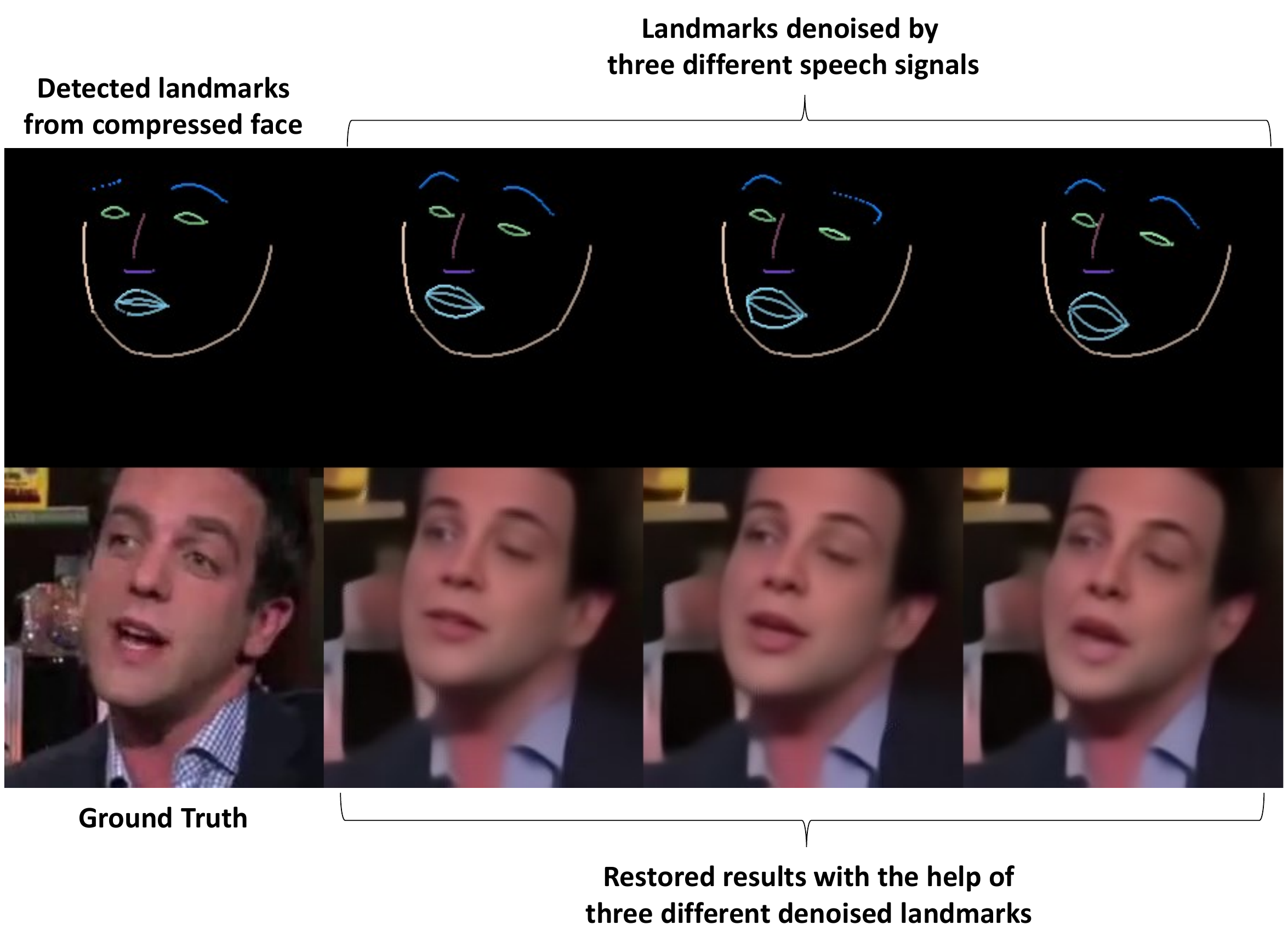}
  \caption{Visualizations of the influence of speech signals on facial landmarks.}
  \label{influence}
\end{figure*}

\subsection{Ablation studies}
In this subsection, we test various ablations of our full architecture to evaluate the effects of each component of the proposed MDVD-Net. We provide both quantitative and qualitative results on various ablations.
Firstly, we build a baseline that only contains the video feature extraction branch and the reconstruction module, then evaluate it using the Obama and VoxCeleb2 datasets which are compressed at 60kbps.  The performance of our baseline is shown in the first row of Table~\ref{ablation}. As expected, the performance of our baseline is comparable to EDVR as they have similar structure and complexity.

\textbf{Ablation of motion vectors.}
The next is to evaluate the impact of using motion vectors to guide the frame alignment. As shown in the second row of Table ~\ref{ablation}, introducing motion vectors for frame alignment can increase the average PSNR of the restored videos by 0.08dB in Obama dataset and 0.03dB in VoxCeleb2 dataset.

\textbf{Ablation of speech signal.}
To evaluate the benefit of using speech in face video restoration, we add the voice feature extraction branch to the baseline network and fuse the voice and video features with a few convolutional layers.  As shown in the third row of Table~\ref{ablation}, the accompanying voices can improve the quality of restored faces by 0.27dB in Obama dataset and 0.16dB in VoxCeleb2 dataset.

\textbf{Ablation of facial landmark correction.}
To assess the impact of speech-aided landmark correction module, we compare the system performances with and without the landmark correction module; the results are reported in the fourth and fifth rows of Table~\ref{ablation}.  It can be seen that after being corrected by speech, the refined facial landmarks can bring appreciable gains on the PSNR of the restored talking head videos.


\textbf{Ablation of spatial attention fusion.}
Also in Table~\ref{ablation}, we compare the restoration performances with and without the proposed spatial attention fusion module.  As shown, by aggregating the speech, video and facial landmark features with attention,
the average PSNR increases by about 0.11dB in Obama dataset and 0.08dB in VoxCeleb2 dataset.


\textbf{Ablation of video codec information.}
We finally evaluate the advantages of exploiting video codec information in deep decompression.
As shown in the last row of Table~\ref{ablation}, by incorporating the video codec information into the network, restoration performance increases by 0.18dB in Obama dataset and 0.11dB in VoxCeleb2 dataset, respectively.

We also provide the visual comparisons of ablation studies on the VoxCeleb2 dataset in Fig.~\ref{abl_eva}.
It can be seen that with the incorporating of various prior information from different modalities, the restored faces have more precise facial shape, sharper lips and clearer muscle contours.
Please refer to the supplementary material for more visual comparisons.

\textbf{PSNR gains of the upper and the lower part of face. }
To further evaluate whether the speech prior benefits only the lips, we analyze the PSNR gains of the upper and the lower part of the face separately in the ablation study of speech signals.
As tabulated in Table~\ref{lu}, the upper and the lower part of the face both achieves PSNR gains when the speech signals are introduced, indicating that the speech prior can benefits both the lips and eyes.
But the performance gain on the lower part of the face is significantly higher than the upper part, indicating that the speech prior contributes more to the reconstruction of the lips than to the eyes.

\begin{table}[t]
  \centering
  \caption{PSNR gains of the upper and the lower part of face in the ablation study of speech signals.
  \textbf{SPB}: speech processing branch;
  \textbf{FLWC}: facial landmark with correction}
  \label{lu}
  \renewcommand\arraystretch{1.4}
  \begin{tabular}{ccccc}
  \hline
   & \multicolumn{2}{c}{Obama} & \multicolumn{2}{c}{VoxCeleb2} \\
  \hline
   & SPB & SPB+FLWC & SPB & SPB+FLWC \\
  \hline
  Upper part & 0.16 & 0.28 & 0.11 & 0.18  \\
  Lower part & 0.38 & 0.66 & 0.21 & 0.36  \\
  \hline
  \end{tabular}
\end{table}

\subsection{How speech signals influence facial landmarks}
Here we visualize and discuss how speech signal influence the prediction of facial landmarks.
In Fig.~\ref{influence}, we show the noisy landmarks detected from the compressed face and the corrected landmarks which are denoised by three different speech signals. In addition, we further use the three denoised landmarks to help restoring compressed face images and exhibit the three different restored results. We can see that different speech signals can control the shape of the predicted landmarks of the mouths/lips, and furthermore, it will affect the shape of the mouths/lips in the restored face images.  

\subsection{Computational complexity}
We evaluate the test speed of the proposed approach and other competing methods
using a computer equipped with a CPU of Intel i7-8700 3.20GHz and a GPU of GeForce GTX 2080 Ti.
Specifically, we measure the average frame per second (fps), when testing
video sequences in VoxCeleb2 dataset.
The results averaged over sequences are reported in Table~\ref{speed}. As shown in this table,
the proposed MDVD method is slightly slower than the existing state-of-the-art method EDVR, but the former
beats all the competing methods in terms of PSNR by a large margin.

\begin{table}[t]
  \centering
  \caption{Test speed (frame per second, FPS) on GPU for $224\times224$ video sequences.}
  \label{speed}
  \renewcommand\arraystretch{1.4}
  \begin{tabular}{ccccccc}
  \hline
  DSCNN & DKFN & MFQE & EDVR & MDVD \\
  \hline
  42.86 & 35.62 & 28.75 & 18.32 & 15.52 \\
  \hline
  \end{tabular}
\end{table}

\subsection{Generalization ability}
To evaluate the generalization ability of the proposed MDVD method, we conduct the following additional experiments: Evaluating the trained models on the videos which are out of the quality range used in the training set. Specifically, the bitrates used for creating the training set are from 60kbps to 120kbps, so we apply the trained models on the compressed videos with the bitrates of 40kbps and 160kpbs, to test the generalization ability of the proposed MDVD method. Table~\ref{gener} shows the generalization results of the competing methods.
It can be seen that even tested on the videos which are beyond the distribution of the training set, the proposed MDVD-Net still achieves the significant performance gains compared to the existing state-of-the-art methods.
\begin{table}[t]
  \centering
  \caption{Performance of the competing methods evaluated on the videos which are beyond the quality range in training.}
  \label{gener}
  \renewcommand\arraystretch{1.4}
  \begin{tabular}{ccccccc}
  \hline
  Bitrate & H.264 & DSCNN & DKFN & MFQE & EDVR & MDVD \\
  \hline
  40kbps & 23.52 & 23.65 & 23.96 & 24.27 & 24.68 & 25.02 \\
  \hline
  160kbps & 32.41 & 32.21 & 32.45 & 32.87 & 33.21 & 33.52 \\
  \hline
  \end{tabular}
\end{table}

\subsection{Application on videos without face alignment}

Our method works the best if the input face video is aligned by landmark points (e.g.,  the voxceleb2 dataset).  For real-world applications in which face videos are not aligned, one can always first align faces through video frames and then apply the proposed MDVD method on these aligned faces to enhance the face image quality; afterwards the inverse alignment can be performed to convert these aligned and enhanced faces back to the original state.  As common face alignment methods are based on affine transformations, their inverse transformations are straightforward. 

\section{Conclusion and Future Work}
We propose, implement and evaluate a novel DCNN system for restoring highly compressed videos of talking heads.
The key innovation is a new DCNN architecture that can incorporate and profit from the known priors of different modalities to repair compression defects in the face region.
We also embed into our network the structural codec information in the video compression standards and introduce a back projection module in the network to further improve the restoration. Experiments show that the proposed MDVD-net outperforms existing methods appreciably.

There may exist two directions of future research.
(1) It will be interesting to investigate if the stereophonic sound may further improve the restoration of compressed face videos.
(2) The proposed method is incapable to predict the direction in which the speaker looks. To remove this limitation, eye gaze can be detected by the encoder and transmitted to the decoder for correct reconstruction of the gaze direction.


\ifCLASSOPTIONcaptionsoff
  \newpage
\fi



%




\bibliographystyle{IEEEtran}
\bibliography{mdvd}

%

\begin{IEEEbiography}
  [{\includegraphics[width=1in,clip,keepaspectratio]{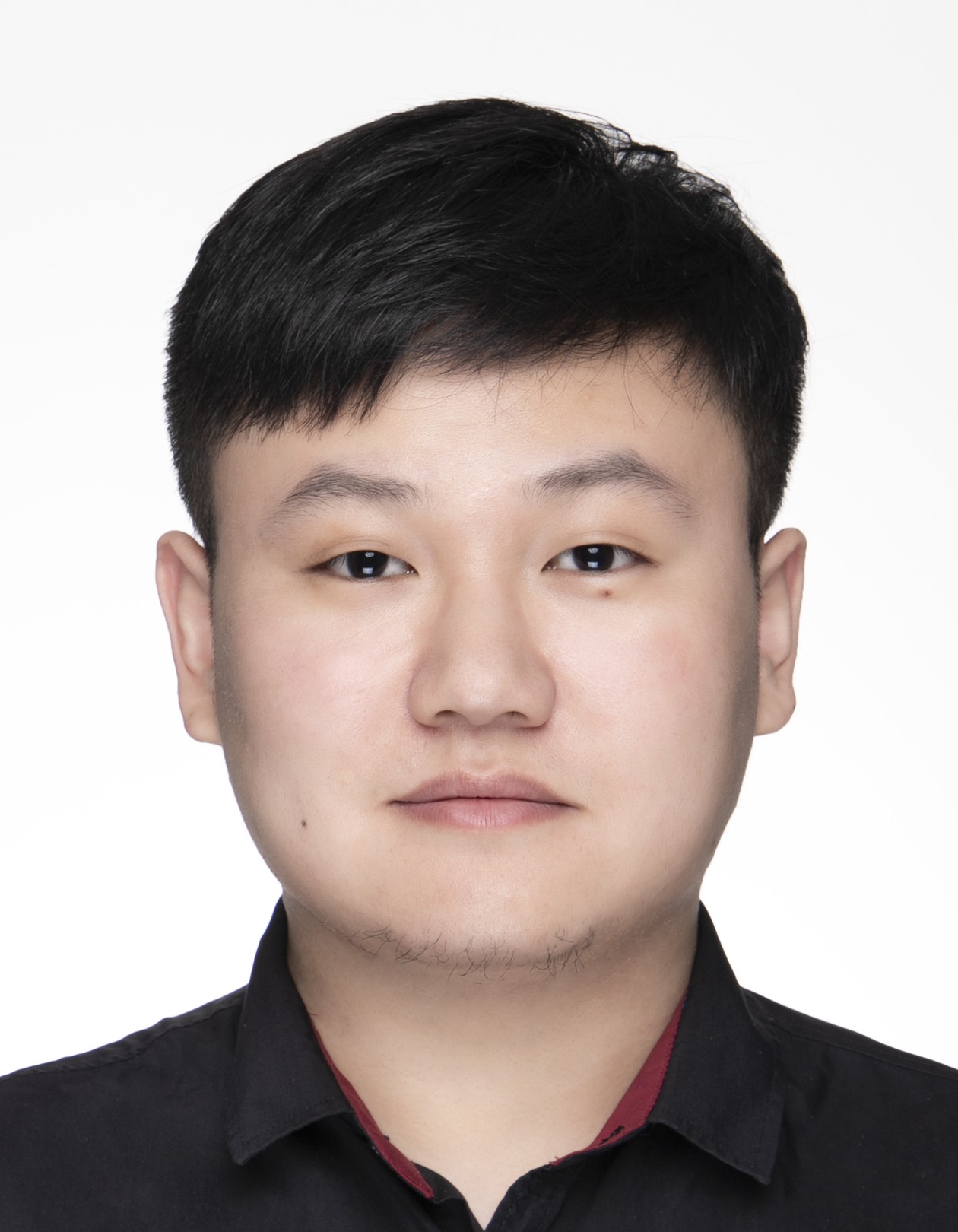}}]{Xi Zhang}
  received the B.Sc. degree in mathematics and physics basic science
  from University of Electronic Science and Technology of China, Chengdu, China, in 2015.
  He is currently pursuing the Ph.D. degree with the Department of Electronic Engineering,
  Shanghai Jiao Tong University, Shanghai, China.
  He is also a visiting Ph.D. student with the 
  Department of Electrical and Computer Engineering, 
  McMaster University, Hamilton, ON, Canada.
  His research interests include image processing, data compression, 
  cognitive computing and visual reasoning.
  \end{IEEEbiography}

  \begin{IEEEbiography}
  [{\includegraphics[width=1in,clip,keepaspectratio]{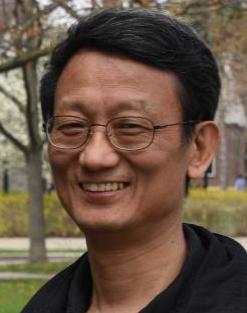}}]{Xiaolin Wu}
  (Fellow, IEEE) received the B.Sc. degree in computer science
  from Wuhan University, China, in 1982, and the Ph.D. degree in computer
  science from the University of Calgary, Canada, in 1988. He started his
  academic career in 1988. He was a Faculty Member with Western University,
  Canada, and New York Polytechnic University (NYU-Poly), USA. He is
  currently with McMaster University, Canada, where he is a Distinguished
  Engineering Professor and holds an NSERC Senior Industrial Research Chair.
  His research interests include image processing, data compression, 
  digital multimedia, low-level vision, and network-aware visual communication.
  He has authored or coauthored more than 300 research articles and holds
  four patents in these fields. He served on technical committees of many IEEE
  international conferences/workshops on image processing, multimedia, data
  compression, and information theory. He was a past Associated Editor of
  IEEE TRANSACTIONS ON MULTIMEDIA. He is also an Associated Editor of
  IEEE TRANSACTIONS ON IMAGE PROCESSING.
  \end{IEEEbiography}








\end{document}